\documentclass[sigconf]{acmart}
\usepackage[linesnumbered,boxed,ruled,commentsnumbered]{algorithm2e}
\usepackage{graphicx}
\usepackage{amsmath}
\usepackage{booktabs} 
\usepackage{color}
\definecolor{ao}{rgb}{0.0, 0.5, 0.0}
\usepackage{comment}
\usepackage{booktabs}
\usepackage{amsfonts}
\usepackage{amsmath}

\usepackage{amssymb}
\usepackage{makecell}
\usepackage{subcaption}
\usepackage{multirow}
\usepackage{pbox}
\usepackage{hhline}
\usepackage{graphicx}
\usepackage[flushleft]{threeparttable}
\usepackage{svg}
\usepackage{courier}
\usepackage{xspace}
\usepackage{epstopdf}
\usepackage{stfloats}
\usepackage{array}
\usepackage{xcolor,colortbl}
\colorlet{shadecolor}{gray!20}

\usepackage{mleftright}

\usepackage[colorinlistoftodos]{todonotes}

\usepackage{bm}
\usepackage{bbm}

\usepackage[aboveskip=1pt]{subcaption}
\usepackage[skip=0.85\baselineskip]{caption}
\usepackage{makecell}

\AtBeginDocument{%
  \providecommand\BibTeX{{%
    \normalfont B\kern-0.5em{\scshape i\kern-0.25em b}\kern-0.8em\TeX}}}

\copyrightyear{2022}
\acmYear{2022}
\setcopyright{rightsretained}

\acmConference[KDD '22]{Proceedings of the 28th ACM SIGKDD Conference on Knowledge Discovery and Data Mining}{August 14--18, 2022}{Washington, DC, USA}
\acmBooktitle{Proceedings of the 28th ACM SIGKDD Conference on Knowledge Discovery and Data Mining (KDD '22), August 14--18, 2022, Washington, DC, USA}
\acmISBN{978-1-4503-9385-0/22/08}
\acmDOI{10.1145/3534678.3539350}

\usepackage{etoolbox}
\makeatletter
\patchcmd{\maketitle}{\@copyrightpermission}{
   \begin{minipage}{0.3\columnwidth}
     \href{https://creativecommons.org/licenses/by/4.0/}{\includegraphics[width=0.90\textwidth]{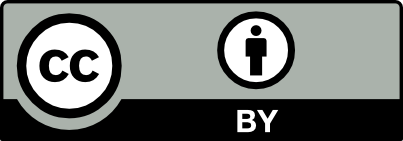}}
   \end{minipage}\hfill
   \begin{minipage}{0.7\columnwidth}
     \href{https://creativecommons.org/licenses/by/4.0/}{This work is licensed under a Creative Commons Attribution International 4.0 License.}
   \end{minipage}
  
   \vspace{5pt}
}{}{}

\makeatother

\begin{document}

\title{Dual-Geometric Space Embedding Model for Two-View Knowledge Graphs}


\author{Roshni G. Iyer}
\affiliation{%
  \institution{University of California, Los Angeles}
  \city{Los Angeles}
  \state{CA}
  \country{USA}
}
\email{roshnigiyer@cs.ucla.edu}

\author{Yunsheng Bai}
\affiliation{%
  \institution{University of California, Los Angeles}
  \city{Los Angeles}
  \state{CA}
  \country{USA}
 }
\email{yba@ucla.edu}

\author{Wei Wang}
\affiliation{%
  \institution{University of California, Los Angeles}
  \city{Los Angeles}
  \state{CA}
  \country{USA}
 }
\email{weiwang@cs.ucla.edu}

\author{Yizhou Sun}
\affiliation{%
  \institution{University of California, Los Angeles}
  \city{Los Angeles}
  \state{CA}
  \country{USA}
}
\email{yzsun@cs.ucla.edu}





\renewcommand{\shortauthors}{Roshni G. Iyer et al.}
\newcommand{\ourmodel}{\textbf{\textsc{DGS}}\xspace}
\newcommand{\transe}{\textbf{\textsc{TransE}}\xspace}
\newcommand{\distmult}{\textbf{\textsc{DistMult}}\xspace}
\newcommand{\complex}{\textbf{\textsc{ComplEx}}\xspace}
\newcommand{\rotate}{\textbf{\textsc{RotatE}}\xspace}
\newcommand{\joie}{\textbf{\textsc{JOIE}}\xspace}
\newcommand{\hyperkg}{\textbf{\textsc{HyperKG}}\xspace}
\newcommand{\hake}{\textbf{\textsc{HAKE}}\xspace}
\newcommand{\cone}{\textbf{\textsc{ConE}}\xspace}
\newcommand{\rehf}{\textbf{\textsc{RefH}}\xspace}
\newcommand{\roth}{\textbf{\textsc{RotH}}\xspace}
\newcommand{\atth}{\textbf{\textsc{AttH}}\xspace}
\newcommand{\hgcn}{\textbf{\textsc{HGCN}}\xspace}
\newcommand{\hyperka}{\textbf{\textsc{HyperKA}}\xspace}
\newcommand{\mtwognn}{\textbf{\textsc{\textrm{M$^{2}$GNN}}}\xspace}
\newcommand{\murp}{\textbf{\textsc{MuRP}}\xspace}

\newcommand\norm[1]{\lVert#1\rVert}

\begin{abstract}
Two-view knowledge graphs (KGs) jointly represent two components: an ontology view for abstract and commonsense concepts, and an instance view for specific entities
that are instantiated from ontological concepts. As such, these KGs contain heterogeneous structures that are hierarchical, from the ontology-view, and cyclical, from the instance-view. Despite these various structures in KGs, most recent works on embedding KGs assume that the entire KG belongs to only one of the two views but not both simultaneously. For works that seek to put both views of the KG together, the instance and ontology views are assumed to belong to the same geometric space, such as all nodes embedded in the same Euclidean space or non-Euclidean product space, an assumption no longer reasonable for two-view KGs where different portions of the graph exhibit different structures. To address this issue, we define and construct a dual-geometric space embedding model (\MakeUppercase{\ourmodel}) that models two-view KGs using a complex non-Euclidean geometric space, by embedding different portions of the KG in different geometric spaces. \ourmodel utilizes the spherical space, hyperbolic space, and their intersecting space 
in a unified framework for learning embeddings. Furthermore, for the spherical space, we propose novel closed spherical space operators that directly operate in the spherical space without the need for mapping to an approximate tangent space. Experiments on public datasets show that \ourmodel significantly outperforms previous state-of-the-art baseline models on KG completion tasks,
demonstrating its ability to better model heterogeneous structures in KGs. 
\end{abstract}

\ccsdesc[500]{Computing methodologies~Knowledge representation
and reasoning}
\ccsdesc{Computing methodologies~non-Euclidean geometric space embedding}

\keywords{Knowledge Graph Embeddings; Non-Euclidean Geometry}


\maketitle

\vspace{-2mm}
\section{Introduction}
\begin{figure*}[h]
    \centering
    \includegraphics[width=0.85\linewidth]{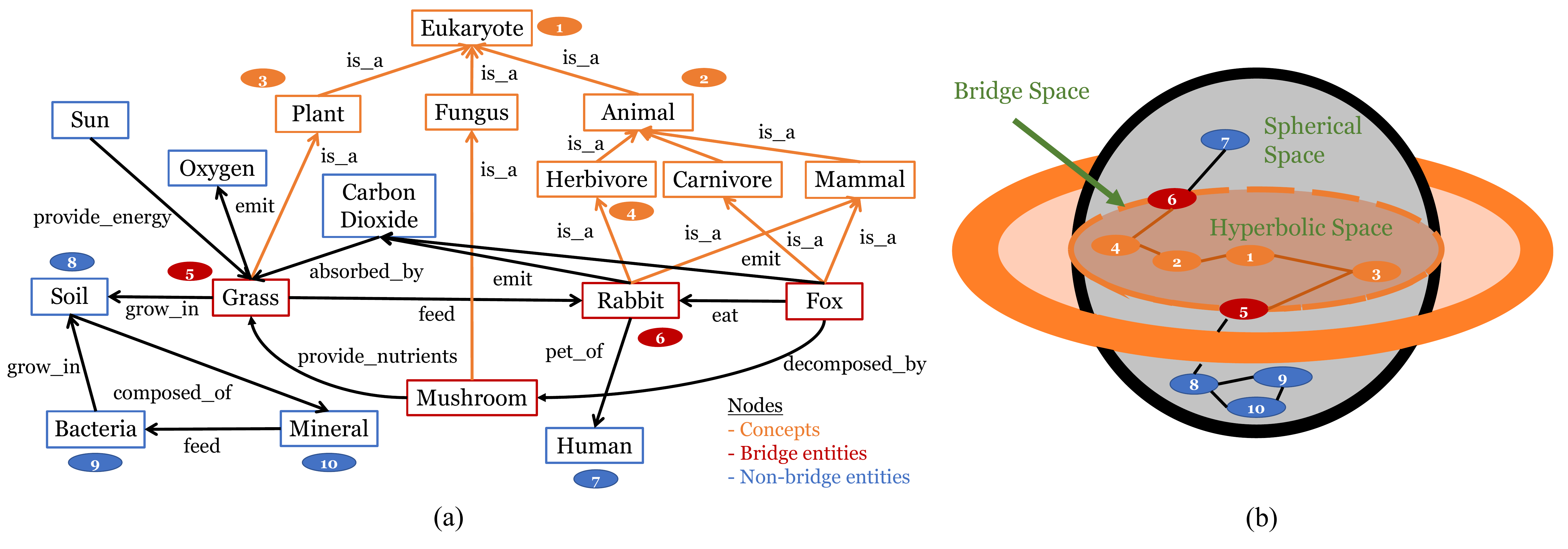}
    \caption{\textmd{Two-View KG visualization and embedding space. (a) Example two-view KG. (b) Corresponding \ourmodel representation.}
    }
    \label{fig:intro-fig}
    \vspace{-3mm}
\end{figure*}

Knowledge graphs (KGs) are essential data structures that have been shown to improve several semantic applications, including semantic search~\cite{kg-sem-parsing}, question answering~\cite{kg-qa, qa-gnn}, and recommender systems~\cite{kg-rec-system}. Two-view knowledge graphs~\cite{general-kg} consist of both (1) an instance-view component, containing entity-entity relations, 
and (2) an ontology-view component, containing concept-concept and entity-concept relations. 
In addition, there are nodes involved in both entity-concept relations and entity-entity relations, which we refer to as \textit{bridge nodes}, e.g., nodes 5 and 6 in Figure~\ref{fig:intro-fig}(a). 
Bridge nodes provide the connections to bridge both KG views. The instance-view component contains data that form cyclical structures, such as nodes 8-10 in Figure~\ref{fig:intro-fig}(a). The ontology-view component contains data that form hierarchical structures. As such, the two-view KG contains intrinsic heterogeneous structures.

A key challenge with KGs is that they are often highly incomplete, making KG completion an important area of investigation. In recent years, several research works have focused on improving knowledge graph embeddings (KGE) to address this task~\cite{transe, distmult}. The idea behind embedding methods is to map entities and relations to a latent low-dimensional vector space while preserving the semantics and inherent structures in the KG. However, most embedding methods, including graph neural network (GNN)-based~\cite{atth, hyperka, hgcn, ba-gnn} and non-GNN-based methods~\cite{hyperkg, complex}, are limited to operating on KGs belonging to either the instance-view or the ontology-view but not both. Specifically, they either model entities and relations 
in the zero-curvature Euclidean space~\cite{rotatE}, omitting both cyclic and hierarchical structures, or in the hyperbolic space~\cite{cone, hake}, omitting cyclic structures and treating relations as hierarchical.

For works that seek to put both views of the KG together, the instance and ontology views are assumed to belong to the same space, an assumption no longer reasonable for two-view KGs. Specifically, all KG relations are modeled using the Euclidean space~\cite{joie}, or are all modeled using a product space, being the Riemannian product manifold of the Euclidean, spherical and hyperbolic spaces~\cite{m2gnn}. However, our goal in modeling two-view KGs 
is to use a unified framework to embed hierarchical structures of the KG in the hyperbolic space, cyclic structures of the KG in the spherical space, and nodes involved in both structures  in a unified intersection space.

To address the above challenges, in this paper, we present the Dual-Geometric Space embedding model (\MakeUppercase{\ourmodel}) for two-view KGs. To summarize, our work makes the following contributions:
\begin{itemize}
    \item We formulate the problem of modeling a two-view KG in complex non-Euclidean geometric space for KG completion. To our knowledge, we are the first to model two-view KGs in distinct non-Euclidean spaces using a unifying framework, e.g., different views belong to different geometric spaces.
    \vspace{0.5mm}
    
    \item We propose to model instance-view entities in the spherical space for solely cyclical relations, and ontology-view concepts in the hyperbolic space for solely hierarchical relations, and bridge entities involved in both cyclic/hierarchical structures in a specially designed intersection bridge space. 
    \vspace{0.5mm}
    
    \item To the best of our knowledge we are also the first to design closed spherical space operations, to directly operate in the spherical space without mapping to an external approximation space, e.g., the tangent Euclidean space. 
    \vspace{0.5mm}

    \item We investigate seven variant and ablation models of \ourmodel and evaluate these models on two KG completion tasks.
    Extensive experiments 
    demonstrate the effectiveness of \ourmodel in populating knowledge in two-view KGs. Further, our model significantly outperforms its single non-Euclidean and Euclidean geometric space counterparts including the product space, and existing state-of-the-art graph neural network (GNN) embedding methods.
    \vspace{0.5mm}
    
\end{itemize}

\vspace{-2mm}
\section{Preliminary and Related Work}
Here, we formalize two-view KGs, which jointly model the entities and concepts in the instance and ontology views, and discuss the achievement of primary state-of-the-art models for two-view KGs.

\vspace{-1mm}
\subsection{Problem Formulation}
A two-view KG, $G$, consists of nodes and edges, such that nodes denote the set of entities $E$, or set of concepts, $C$. Edges denote relations where $R_{I}$ are the set of entity-entity relations, $R_{O}$ are the set of concept-concept relations, and $R_{IO}$ are the set of entity-concept relations. The two-view KG consists of an instance-view component containing entities with relations of $R_{I}$, and an ontology-view component containing concepts and bridge entities with relations of $R_{O}$ and $R_{IO}$. We denote a subset of entities in $E$ to be bridge entities, that communicate between both instance and ontology views. These bridge entities associate with both $R_{I}$ and $R_{IO}$ relations. Relations in $R_{IO}$ are entity-concept relations,  which are similar to concept-concept relations 
such as ``\textit{is\_a}''. $E$ and $C$ as well as $R_{I}$, $R_{\mathcal{O}}$ and $R_{I\mathcal{O}}$ are each disjoint sets. We denote $e_{i} \in E$ to be the $i$-th entity, $\boldsymbol{h}_{e_{i}}$ to be entity $i$'s representation in the Cartesian coordinate system, and $r_{Ik} \in R_{I}$ to be the $k$-th relation in the instance-view component.
Without loss of generality (WLOG), we also model the ontology-view component where $c_{i} \in C$ is the $i$-th concept, $\boldsymbol{h}_{c_{i}}$ is concept $i$'s representation in the Cartesian coordinate system, and $r_{\mathcal{O}k} \in R_{\mathcal{O}}$ is the $k$-th relation in the ontology-view component.
Therefore, the instance-view can be considered as a set of triples in the form  
$(e_{i}, {r_{Ik}}, e_{j}) \in E \times R_{I} \times E$, which may be any real-world instance associations and are often cyclical in structure. Likewise, the ontology-view can be represented by a set of triples 
$(c_{i}, {r_{\mathcal{O}k}}, {c_{j}}) \in C \times R_{\mathcal{O}} \times C$, which contain hierarchical associations such as \textit{is\_a}.
Here we use
Figure~\ref{fig:intro-fig}(a) to illustrate the aforementioned problem formulation. For example, (\textit{Soil, composed\_of, Mineral}) and (\textit{Fungus, is\_a, Eukaryote}) are instance-view and ontology-view triples respectively. Further, nodes of \textit{Bacteria-Soil-Mineral} is an example of a cycle between non-bridge entities. 

A bridge entity may be involved with two types of triples:\\
$({e_{i}}, {r_{I\mathcal{O}k}}, {c_{j}})$ $\in E \times R_{I\mathcal{O}} \times C$ for (hierarchical) entity-concept relations and 
$({e_{i}}, {r_{Ik}}, {e_{j}})$ $\in E \times R_{I} \times E$ for (cyclical) entity-entity relations. 
For example,  \textit{Rabbit} is a bridge entity in Figure~\ref{fig:intro-fig}(a).  The triple (\textit{Rabbit, emit, Carbon Dioxide}) represents an entity-entity relation, and the triple (\textit{Rabbit, is\_a, Herbivore}) represents an entity-concept relation. 

The objective of our research is to learn KG embeddings of nodes and relations in the KG, such that we seamlessly unify multiple curvature geometric spaces to better capture the contextual information and heterogeneous structures in the KGs. We evaluate the quality of the learned embeddings on the KG tasks of triple completion and entity typing, described in Section~\ref{sec:exp}.

\vspace{-3.6mm}
\subsection{Non-Euclidean Geometric Spaces}
\label{subsec:non-euc-space}
In this section, we describe the various properties of non-Euclidean geometric spaces, which are curved spaces unlike the zero-curvature Euclidean space. The textbook~\cite{reim-geom} provides more details. Geometric spaces of Euclidean ($\mathbb{E}^{d}$), spherical ($\mathbb{S}^{d}$), and hyperbolic ($\mathbb{H}^{d}$) spaces belong to Riemannian manifolds ($\boldsymbol{M}^{d}$), such that each point $\boldsymbol{a} \in \boldsymbol{M}^{d}$ has a corresponding tangent space, $(T_{\boldsymbol{a}}\boldsymbol{M}^{d})^{d}$, that
approximates $\boldsymbol{M}^{d}$ around $\boldsymbol{a}$. Further, each Riemmanian manifold, $\boldsymbol{M}^{d}$, is associated with a Riemanian metric $dist$ that defines the geodesic distance of two points on the manifold and the curvature $K$ of the space. In the spherical space, curvature $K_{S} > 0$, suitable for capturing cyclical structures~\cite{zhang2021switch}, while in the hyperbolic space, curvature $K_{H} < 0$, suitable for capturing hierarchical structures~\cite{hyperbolic-dist}. Widely used models on the hyperbolic space include the Poincaré ball model~\cite{poincare-orig}, the Lorentz~\cite{lorentz-orig}
model, and the Klein model~\cite{lorentz-orig}. As the these three models are both isometric and isomorphic to one another, WLOG we utilize the Poincaré ball model in our work. 

\paragraph{\normalfont{\textbf{Non-Euclidean Space Optimization}}}
\ourmodel utilizes Riemannian optimization for updating entity and relational embeddings because Euclidean space optimization methods, such as SGD, provide the update direction of the Euclidean gradient to be in non-curvature space. This does not align with parameters in our model that must be updated in positive or negative curvature spaces. For parameter learning, \ourmodel uses RSGD~\cite{rsgd}, whose update function is denoted below, where $(T_{\boldsymbol{a}}M^{d})^{d}$ denotes the tangent Euclidean space of $\boldsymbol{a} \in M^{d}$, $\nabla_{R} \in (T_{\boldsymbol{a}}M^{d})^{d}$ denotes the Riemannian gradient of loss function $L(\boldsymbol{a})$, $\mathcal{R}_{\boldsymbol{a}_{t}}$ denotes retraction onto $M^{d}$, or non-Euclidean space at $\boldsymbol{a}$, and $\eta_{t}$ denotes learning rate at time $t$:
\begin{equation*}
    \boldsymbol{a_{t+1}} = \mathcal{R}_{\boldsymbol{a}_{t}}(-\eta_{t}\nabla_{R}L(\boldsymbol{a}_{t}))
\end{equation*}

The retraction operator, $\mathcal{R}(\cdot)$ involves mapping between spaces. For non-Euclidean spaces, the retraction is generally performed between the non-Euclidean space and approximate tangent Euclidean space using logarithmic and exponential mapping functions as follows, where  $\mathrm{log}_{0}(\boldsymbol{h}_{e_{i}}^{B})$ is a logarithmic map at center $\boldsymbol{0}$ from the hyperbolic space to Euclidean tangent space, and $\mathrm{exp}_{0}(\boldsymbol{h}_{e_{i}}^{B})$ is an exponential map at center $\boldsymbol{0}$ from the Euclidean tangent space to hyperbolic space of $\boldsymbol{a}$.
\begin{equation}
    \mathrm{log}_{0}(\boldsymbol{a}) = \mathrm{tanh}^{-1}(i \cdot \norm{\boldsymbol{a}})\frac{\boldsymbol{a}}{i \cdot \norm{\boldsymbol{a}}}
\end{equation}
\begin{equation}
    \mathrm{exp}_{0}(\boldsymbol{a}) = \mathrm{tanh}(i \cdot \norm{\boldsymbol{a}})\frac{\boldsymbol{a}}{i \cdot \norm{\boldsymbol{a}}}
\end{equation}

\subsection{Two-View KG Models}
\label{gen-kg-mod}
In this section, we describe the models that are utilized for two-view KGs, which consider the problem setting of modeling ontological and instance views. To address the challenges of these models, we propose \ourmodel in Section~\ref{dgs-arch}. 

\paragraph{\normalfont{\joie}}
The \joie~\cite{joie} model is an embedding model that considers the problem of two-view KGs. However, \joie models all triples in the same zero-curvature Euclidean space, omitting the hierarchical and cyclical structures of the KG. Further, there is no special consideration of representation for bridge nodes. 

\paragraph{\normalfont{\textbf{Models leveraging product of spaces}}} Models including \\ \mtwognn~\cite{m2gnn} which extends~\cite{gu2018learning} from social networks to the domain of KGs, \cite{sun2021self}, etc. help to address the limitation of \joie by utilizing the hyperbolic and spherical spaces for representing triples. However, they treat all triples in the KG to be in the same non-Euclidean product space, formed by a Riemannian product of the Euclidean, spherical, and hyperbolic spaces. Thus, the embedding space of hierarchical triples is not distinguished from the embedding space of cyclic triples, and further there is also no special consideration of representation for bridge nodes.

\section{\ourmodel Architecture}
\label{dgs-arch}
This section describes our proposed dual-geometric space embedding model (\MakeUppercase{\ourmodel}), which jointly embeds entities and concepts in a knowledge graph. \MakeUppercase{\ourmodel} embeds different link types of the two-view KG in different non-Euclidean manifold spaces to capture the inherent heterogeneous structures of the KG, as described below. We also design general procedures, which are detailed in the Supplements, for efficiently converting arbitrary $d$-dimensional embeddings between the polar and Cartesian coordinate representations, which are utilized by sub-models in \ourmodel. Source code is available at \href{https://github.com/roshnigiyer/dgs}{\texttt{https://github.com/roshnigiyer/dgs}}.

\subsection{Modeling}
\label{subsec:dgs-modeling}

The key questions in the modeling process are in how to: (1) design or adopt an appropriate embedding space for nodes, (2) define the representation of entity/concept and relation in that embedding space, and (3) define the KG embedding model and loss function of \ourmodel. Our framework enables each of entities and concepts to be influenced by bridge nodes, and simultaneously bridge nodes to be informed by entities and concepts. In this way, \ourmodel exhibits a unified framework to jointly model two-views of KGs.

\ourmodel models nodes in the instance view as points in a spherical space $\mathbb{S}^{d}$ with fixed norm space $w^{S}$, and nodes in the ontology view as points in a hyperbolic space $\mathbb{H}^{d}$ with learnable norm space $w_{\boldsymbol{h}_{c_{i}}^{H}}$ per concept. The bridge nodes lie in the intersection of the two, which is a submanifold intersection space, called $\mathbb{B}^{d}$, shown as the dotted circle in Figure~\ref{fig:intro-fig}(b). $\mathbb{B}^{d}$ contains the same fixed norm space as the spherical space $w^{S}$. For modeling compatibility, we set the degrees of freedom of $\mathbb{S}^{d}$, $\mathbb{H}^{d}$, and $\mathbb{B}^{d}$ to $d-1$, $d-1$, and $d-2$ respectively. $\mathbb{B}^{d}$ has one degree of freedom less than $\mathbb{H}^{d}$ and $\mathbb{S}^{d}$ because it is a submanifold intersection space of both the spherical and hyperbolic spaces. The norm of $\mathbb{B}^{d}$ is $w^{S}$ because that is the intersection norm space of $\mathbb{S}^{d}$ and $\mathbb{H}^{d}$. The concept-specific norm spaces, $w_{\boldsymbol{h}_{c_{i}}^{H}}$, are learnable in order for the hierarchy in the KG to be learned by the embedding. In practice, it can be seen that hierarchical root concepts move towards the center of the hyperbolic space e.g., towards norm 0, shown in Section~\ref{sec:exp}. 

Parameter optimization, detailed in Section~\ref{subsec:mgs-training}, is performed using RSGD, described in Section~\ref{subsec:non-euc-space}, on hinge loss functions which utilize non-Euclidean space geodesic distances for the spherical and hyperbolic spaces respectively. We construct the hinge loss function such that positive triples are scored higher than negative triples within positive margin hyperparameters.

\subsubsection{Modeling Instance View KG in Spherical Space}
\label{subsec:inst-modeling}

\paragraph{\normalfont{\textbf{Representation of entities.}}} We propose to embed the entities from the instance-view on the surface of the spherical ball from the spherical space, $\mathbb{S}^{d}$, in order to better capture the cyclic structures present in this KG view. Entities are represented as points $\boldsymbol{h}_{e_{i}}$ that belong to the surface of the spherical ball in $\mathbb{S}^{d}$ as shown in Figure~\ref{fig:intro-fig}(b). Formally, $\mathbb{S}^{d} = \{ \boldsymbol{h}_{e_{i}}^{S} \in \mathbb{R}^{d} \big| \norm{\boldsymbol{h}_{e_{i}}^{S}} = w^{S} \}, w^{S} \in [0, 1)$, is the $d$-dimensional $w^{S}$-norm ball, where $\norm{\cdot}$ is the Euclidean norm.


For an entity $\boldsymbol{h}_{e_{i}}^{S}$, we propose to model the relation applied to the entity as a rotation operation, and therefore represent the relation $r_{Ik}$ as a vector of angles $\boldsymbol{\theta}_{r_{Ik}}^{S} \in [0, 2\pi)^{d-1}$.
Below, we show our proposed vector transformation procedure to model the relation as rotation of the head entity, $f_{\mathrm{rot}}: \boldsymbol{h}_{e_{i}}^{S, new} =  f_{\mathrm{rot}}(\boldsymbol{h}_{e_{i}}^{S}, \boldsymbol{\theta}_{r_{Ik}}^{S})$, and prove that the operation is closed in the spherical space. $f_{\mathrm{rot}}(\cdot)$ eliminates the need to project representations to the tangent Euclidean space in order to perform standard transformations. To the best of our knowledge, we are the first to propose a closed vector transformation procedure that operates directly in the spherical space.

\paragraph{\normalfont{\textbf{Spherical Space Vector Transformation Procedure: $f_{\mathrm{rot}}(\cdot)$.}}} This section describes our proposed vector transformation procedure, $f_{\mathrm{rot}}$, for the spherical space, which directly performs operations using properties, such as positive curvature, in the spherical space. This section also outlines the proof of the closedness property of the transformation procedure. For computational efficiency, we also extend the vector transformation procedure to the batch version, the batch vector transformation procedure, which is detailed in the Supplements. $f_{\mathrm{rot}}$ takes as input an entity embedding and relation operator, which it uses to transform the entity embedding: $\boldsymbol{h}_{e_{i}}^{S, new} = f_{\mathrm{rot}}(\boldsymbol{h}_{e_{i}}^{S}, \boldsymbol{\theta}_{r_{Ik}}^{S})$:

\begin{enumerate}
    \item Given $\boldsymbol{h}_{e_{i}}^{S}, \boldsymbol{\theta}_{r_{Ik}}^{S}$, we convert $\boldsymbol{h}_{e_{i}}^{S}$ to the corresponding representation $\boldsymbol{\theta}_{e_{i}}^{S}$ in polar coordinates, with $\mathit{rad}$ denoting the radius which has the value $w^{S}$. Refer to Section~\ref{sec-mcp} for details on the conversion procedure. 
    \begin{equation}
        \boldsymbol{h}_{e_{i}}^{S}: [x_{e_{i},1}^{S}, x_{e_{i},2}^{S}, ..., x_{e_{i},d}^{S}] \rightarrow \boldsymbol{\theta}_{e_{i}}^{S}: [\theta_{e_{i},1}^{S}, \theta_{e_{i},2}^{S}, ..., \theta_{e_{i},d-1}^{S}]; \mathit{rad}=w^{S}
    \end{equation}
    \begin{equation}
        \boldsymbol{\theta}_{r_{Ik}}^{S}: [\theta_{r_{Ik},1}, \theta_{r_{Ik},2}, ..., \theta_{r_{Ik},d-1}]; \mathit{rad}=w^{S}
    \end{equation}
    
    \item Denote $\boldsymbol{z}[l]$ to be the $l$-th entry of $\boldsymbol{z}$ and apply the transformation: 
    \begin{equation}
        (\boldsymbol{\theta}_{e_{i}}^{S} + \boldsymbol{\theta}_{r_{Ik}}^{S})[l] = (\boldsymbol{\theta}_{e_{i}}^{S}[l] + \boldsymbol{\theta}_{r_{Ik}}^{S}[l]) \  \mathrm{mod} \ 2\pi; l \in [1, d-1]
    \end{equation}
    \begin{multline}
        \boldsymbol{\theta}_{e_{i}}^{S, new} = [(\theta_{e_{i},1}^{S} + \theta_{r_{Ik},1}^{S}) \ \mathrm{mod}\  2\pi, ...,\\ (\theta_{e_{i},d-1}^{S} + \theta_{r_{Ik},d-1}^{S})\  \mathrm{mod} \  2\pi]; \mathit{rad} = w^{S}
    \end{multline}
    
    \item Convert from polar coordinates back to Cartesian coordinates. Refer to Section~\ref{sec-mpc} for details on the conversion procedure.
    \begin{equation}
        \boldsymbol{\theta}_{e_{i}}^{S, new} \rightarrow \boldsymbol{h}_{e_{i}}^{S, new}: [x_{e_{i},1}^{S, new}, x_{e_{i},2}^{S, new},..., x_{e_{i},d}^{S, new}]
    \end{equation}
    
\end{enumerate}

\paragraph{\normalfont{\textbf{Theorem.}}} The vector transformation procedure, $f_{\mathrm{rot}}$,  is closed in the spherical space. 

\begin{proof}
The proof is outlined by examining all three steps for $f_{\mathrm{rot}}$ from the transformation procedure, where $\boldsymbol{h}_{e_{i}}$ and $\boldsymbol{\theta}_{e_{i}}$ represent the same point $e_{i}$ in the embedding space by the isomorphic Cartesian and polar coordinate systems respectively.

\begin{enumerate}
    \item The Cartesian coordinate representation is equivalent to the polar coordinate representation of the point, $e_{i}$, under MCP model, detailed in the Supplements. Further, radius $\mathit{rad} = w^{S}$ of the polar representation embeddings lies in the spherical norm space $w^{S}$.   
    
    \item  $\theta \in [0, 2\pi); \mathit{rad} = w^{S}$ defines polar representation embeddings in the spherical space, and \\ $(\boldsymbol{\theta}_{e_{i}}^{S} + \boldsymbol{\theta}_{r_{Ik}}^{S})[l] = (\boldsymbol{\theta}_{e_{i}}^{S}[l] + \boldsymbol{\theta}_{r_{Ik}}^{S}[l]) \ \mathrm{mod} \ 2\pi \in [0, 2\pi) \forall l; \mathit{rad} = w^{S}$. Thus, both angular and radial coordinates are preserved. 
    
    \item The polar coordinate representation is equivalent to the Cartesian coordinate representation of the point, $e_{i}$, under MPC model, detailed in the Supplements. 
    
\end{enumerate}
\vspace{-2mm}
\end{proof}

\vspace{-3.5mm}
\paragraph{\normalfont{\textbf{Instance View Loss Function.}}}
The instance-view model uses a hinge loss function that is maximized for all triples in the instance space $R_{I}$, with positive triples denoted, $\mathrm{pt}_{I} = (\boldsymbol{h}_{e_{i}}^{S}, \boldsymbol{\theta}_{
r_{Ik}}^{S}, \boldsymbol{h}_{e_{j}}^{S})$ with corresponding score $\phi^{S}_{\mathrm{obs}}(\mathrm{pt}_{I})$ and negative triples denoted, $\mathrm{nt}_{I} = (\boldsymbol{h}_{e_{i}}^{S'}, \boldsymbol{\theta}_{
r_{Ik}}^{S}, \boldsymbol{h}_{e_{j}}^{S'})$ with corresponding score $\phi^{S}_{\mathrm{corr}}(\mathrm{nt}_{I})$, and $\gamma^{R_{I}} > 0$ is a positive margin hyperparameter. Specifically, the instance loss function measures the distance between the predicted tail entity and the ground truth.
\begin{equation}
    \phi^{S}_{\mathrm{obs}}(\mathrm{pt}_{I}) = \mathrm{dist}_{S}\big(f_{\mathrm{rot}}(\boldsymbol{h}_{e_{i}}^{S}, \boldsymbol{\theta}_{r_{Ik}}^{S}), \boldsymbol{h}_{e_{j}}^{S}\big)
\end{equation}
\begin{equation}
    \phi^{S}_{\mathrm{corr}}(\mathrm{nt}_{I}) = \mathrm{dist}_{S}\big(f_{\mathrm{rot}}(\boldsymbol{h}_{e_{i}}^{S'}, \boldsymbol{\theta}_{r_{Ik}}^{S}), \boldsymbol{h}_{e_{j}}^{S'}\big)
\end{equation}
\begin{equation}
    L_{\mathrm{inst}}^{R_{I}} = \frac{1}{|R_{I}|} \sum_{(\mathrm{pt_{I}} \in R_{I}) \land (\mathrm{nt_{I}} \notin R_{I})}{\mathrm{max}\big(0, \gamma^{R_{I}} + \phi^{S}_{\mathrm{obs}}(\mathrm{pt}_{I}) - \phi^{S}_{\mathrm{corr}}(\mathrm{nt}_{I})\big)}
\end{equation}
We calculate spherical geodesic distance~\cite{spherical-dist} between points $\boldsymbol{x}^{S}$ and $\boldsymbol{y}^{S}$ on the manifold as follows: 
\begin{equation}
    \mathrm{dist}_{S}(\boldsymbol{x}^{S}, \boldsymbol{y}^{S}) = \mathrm{arccos}((\boldsymbol{x}^{S})^{^T}\boldsymbol{y}^{S})
\end{equation}
$R_{I}$ includes links between all entities including both non-bridge entities and bridge-entities. In this way, the representation of non-bridge entities is influenced by the bridge entities.

\subsubsection{Modeling Ontological View KG in Hyperbolic Space}
\label{subsec:onto-modeling}

\paragraph{\normalfont{\textbf{Representation of concepts.}}} We propose to embed concepts from the ontology-view on the Poincaré disk from the hyperbolic space, $\mathbb{H}^{d}$, in order to better capture the hierarchical structures present in this KG view. Concepts are represented as points, $\boldsymbol{h}_{c_{i}}$ that belong inside the Poincaré disk in $\mathbb{H}^{d}$ as shown in Figure~\ref{fig:intro-fig}(b). Formally, $\mathbb{H}^{d} = \{ \boldsymbol{h}_{c_{i}}^{H} \in \mathbb{R}^{d} \big| \norm{\boldsymbol{h}_{c_{i}}^{H}} = w_{c_{i}}^{H} \}, w_{c_{i}}^{H} \in [0, 1)$, is the $d$-dimensional $w_{c_{i}}^{H}$-norm ball, where $\norm{\cdot}$ is the Euclidean norm. We assume the center of the disk is aligned with the center of the sphere, and for convenience set the last dimension $d$ to 0.


For concepts in the hyperbolic space, $\mathbb{H}^{d}$, any hyperbolic space model for KG can be applied in principle, which we denote as follows with $r_{\mathcal{O}k} \in R_{\mathcal{O}}$ to denote a relation between two concepts:
\begin{equation}
    f_{\mathrm{KGE}}(\mathrm{pt}_{O}) = f_{\mathrm{KGE}}(\boldsymbol{h}_{c_{i}}^{H}, r_{\mathcal{O}k}, \boldsymbol{h}_{c_{j}}^{H})
\end{equation}

We illustrate \murp~\cite{murp} as an example scoring function, which uses the hyperbolic geodesic distance and relies on Möbius addition to model the relation, where $\mathrm{exp}_{0}(\cdot)$ and $\mathrm{log}_{0}(\cdot)$ are defined in
Section~\ref{subsec:non-euc-space},
$\boldsymbol{R} \in \mathbb{R}^{d \times d}$ is a learnable diagonal relation matrix representing the stretch transformation by relation $r_{\mathcal{O}k} \in R_{\mathcal{O}}$ with representation $\boldsymbol{h}_{r_{\mathcal{O}k}}^{H}$, and scalar biases $b_{c_{i}}^{H}, b_{c_{j}}^{H}$ of concepts $c_{i}$ and $c_{j}$:
\begin{equation}
\begin{split}
    f_{\mathrm{MuRP}}(\mathrm{pt}_{O}) & =
    f_{\mathrm{MuRP}}(\boldsymbol{h}_{c_{i}}^{H}, r_{\mathcal{O}k}, \boldsymbol{h}_{c_{j}}^{H})
    \\ & = f_{\mathrm{MuRP}}(\boldsymbol{h}_{c_{i}}^{H}, \boldsymbol{h}_{r_{\mathcal{O}k}}^{H}, \boldsymbol{h}_{c_{j}}^{H})
    \\ & = -\mathrm{dist}_{H}\big(\mathrm{exp}_{0}(\boldsymbol{R}\mathrm{log}_{0}(\boldsymbol{h}_{c_{i}}^{H})\big), \boldsymbol{h}_{c_{j}}^{H} \oplus_{H} \boldsymbol{h}_{r_{\mathcal{O}k}}^{H})^{2} + b_{c_{i}}^{H} + b_{c_{j}}^{H}
\end{split}
\end{equation}


\begin{equation}
    \boldsymbol{x}^{H} \oplus_{H} \boldsymbol{y}^{H} = \frac{(1 + 2\langle\boldsymbol{x}^{H}, \boldsymbol{y}^{H}\rangle + \norm{\boldsymbol{y}^{H}}_{2}^{2})\boldsymbol{x}^{H} + (1 - \norm{\boldsymbol{x}^{H}}_{2}^{2})\boldsymbol{y}^{H}}{1 + 2\langle\boldsymbol{x}^{H},\boldsymbol{y}^{H}\rangle + \norm{\boldsymbol{x}^{H}}_{2}^{2}\norm{\boldsymbol{y}^{H}}_{2}^{2}} 
\end{equation}

We calculate hyperbolic geodesic distance~\cite{hyperbolic-dist} between points $\boldsymbol{x}^{H}$ and $\boldsymbol{y}^{H}$ on the manifold as follows:
\begin{equation}
    \mathrm{dist}_{H}(\boldsymbol{x}^{H}, \boldsymbol{y}^{H}) = \mathrm{arccosh}(1+ 2\frac{\norm{\boldsymbol{x}^{H} - \boldsymbol{y}^{H}}_{2}^{2}}{(1 - \norm{\boldsymbol{x}^{H}}^{2})(1 - \norm{\boldsymbol{y}^{H}}^{2})})
\end{equation}

\paragraph{\normalfont{\textbf{Ontological View Loss Function.}}} The ontology-view model uses a hinge loss function that is maximized for all links between concepts in the ontology space and all links between bridge nodes and concepts,  i.e., $R_{\mathcal{O}} \cup R_{I\mathcal{O}}$, with positive triples denoted, $\mathrm{pt}_{\mathcal{O}} = (\boldsymbol{h}_{c_{i}}^{H}, \boldsymbol{\theta}_{
r_{\mathcal{O}k}}^{H}, \boldsymbol{h}_{c_{j}}^{H}) \cup (\boldsymbol{h}_{e_{i}}^{H}, \boldsymbol{h}_{
r_{I\mathcal{O}k}}, \boldsymbol{h}_{c_{j}}^{H})$ with corresponding score $\phi^{H}_{obs}(\mathrm{pt}_{\mathcal{O}})$ and negative triples denoted, $\mathrm{nt}_{\mathcal{O}} = (\boldsymbol{h}_{c_{i}}^{H'}, \boldsymbol{\theta}_{
r_{\mathcal{O}k}}^{H}, \boldsymbol{h}_{c_{j}}^{H'}) \cup (\boldsymbol{h}_{e_{i}}^{H'}, \boldsymbol{h}_{
r_{I\mathcal{O}k}}, \boldsymbol{h}_{c_{j}}^{H'})$ with corresponding score $\phi^{H}_{corr}(\mathrm{nt}_{\mathcal{O}})$, and $\gamma^{R_{\mathcal{O}} \cup R_{I\mathcal{O}}} > 0$ is a positive margin hyperparameter:
\begin{equation}
    \phi^{H}_{obs}(\mathrm{pt}_{\mathcal{O}}) = f_{\mathrm{KGE}}(\mathrm{pt}_{\mathcal{O}})
\end{equation}
\begin{equation}
    \phi^{H}_{corr}(\mathrm{nt}_{\mathcal{O}}) = f_{\mathrm{KGE}}(\mathrm{nt}_{\mathcal{O}})
\end{equation}
\begin{equation}
\begin{split}
    L_{\mathrm{onto}}^{R_{\mathcal{O}} \cup R_{I\mathcal{O}}} = \frac{1}{|R_{\mathcal{O}} \cup R_{I\mathcal{O}}|} \sum_{(\mathrm{pt_{\mathcal{O}}} \in R_{\mathcal{O}} \cup R_{I\mathcal{O}}) \land (\mathrm{nt_{\mathcal{O}}} \notin R_{\mathcal{O}} \cup R_{I\mathcal{O}})}\\{\mathrm{max}\big(0, \gamma^{R_{\mathcal{O}} \cup R_{I\mathcal{O}}} + \phi^{H}_{corr}(\mathrm{nt}_{\mathcal{O}}) - \phi^{H}_{obs}(\mathrm{pt}_{\mathcal{O}})\big)}
\end{split}
\end{equation}
$R_{O} \cup R_{I\mathcal{O}}$ includes triples composed of concepts and bridge-entities. In this way, the representation of concepts is influenced by the bridge entities, which is described in Section~\ref{subsec:model-bridge}.

\subsubsection{Modeling the Intersection of the Two Spaces.}
\label{subsec:model-bridge}

\paragraph{\normalfont{\textbf{Representation of bridge entities.}}} 

Bridge nodes are entity nodes that bridge the communication between the instance-view and ontology-view components. Bridge nodes are connected to concepts in the graph, formed by link ontology $\boldsymbol{h}_{r_{I\mathcal{O}k}}$ but may also be connected to other entities, formed by link instance $\boldsymbol{\theta}_{r_{\mathcal{I}k}}$. As shown in Figure~\ref{fig:intro-fig}(a), nodes 5 and 6 are bridge nodes that are involved in both cyclic and hierarchical structures through their links to other entities as well as concepts. As such, we propose to embed bridge entities in the intersection space of the Poincare disk and surface of the spherical ball, in order to better capture the heterogeneous KG structures that are associated with these nodes. We refer to the intersection submanifold embedding space as the bridge space, $\mathbb{B}^{d}$, where the representation of these nodes are informed by both $\mathbb{S}^{d}$ and $\mathbb{H}^{d}$ and has one lower degree of freedom. The bridge space can therefore be derived as a sphere in general. Formally, $\mathbb{B}^{d} = \{ \boldsymbol{h}_{e_{i}}^{B} \in \mathbb{R}^{d} \big| \norm{\boldsymbol{h}_{e_{i}}^{B}} = w^{S}\}, w^{S} \in [0, 1)$, is the $d$-dimensional $w^{S}$-norm ball, where $\norm{\cdot}$ is the Euclidean norm, and the value of last dimension $d = 0$. Links associated with bridge nodes are $\boldsymbol{\theta}_{r_{Ik}}, \boldsymbol{h}_{r_{I\mathcal{O}k}} \in [0, 2\pi)^{d-1}$, and operations on bridge nodes, such as geodesic distance and loss functions, happen in either the spherical space or hyperbolic space. 

To ensure compatibility with the connected concept nodes, we map the bridge entities, $\boldsymbol{h}_{e_{i}}^{B}$, to an embedding in the ontology space through a non-linear transformation function, $g_{\boldsymbol{h}_{r_{I\mathcal{O}k}}}(\boldsymbol{h}_{e_{i}}^{B})$, where $\mathrm{AGG}(\cdot)$ denotes an averaging over all relations $k$ in $R_{IO}$. Logarithmic and exponential mapping functions of $\mathrm{log}_{0}(\cdot)$ and $\mathrm{exp}_{0}(\cdot)$ are described in Section~\ref{subsec:non-euc-space}.
\begin{equation}
    g_{\boldsymbol{h}_{r_{I\mathcal{O}k}}}(\boldsymbol{h}_{e_{i}}^{B}) = \textrm{AGG}\Big( \mathrm{proj}_{B}\big(\mathrm{tanh}(\boldsymbol{W}_{\boldsymbol{h}_{r_{IOk}}} \otimes_{H} \boldsymbol{h}_{e_{i}}^{B} \oplus_{H} \boldsymbol{b}_{\boldsymbol{h}_{r_{IOk}}})\big)\Big)
\end{equation}
\begin{equation}
    \mathrm{proj}_{B}(\boldsymbol{z})=\mathrm{proj}_{S}(\boldsymbol{z})
\end{equation}
\begin{equation}
    \boldsymbol{M} \otimes_{H} \boldsymbol{h}_{e_{i}}^{B} = 
    \mathrm{exp}_{0}\big(\boldsymbol{M}\mathrm{log}_{0}(\boldsymbol{h}_{e_{i}}^{B})\big)
\end{equation}
\begin{equation}
    \boldsymbol{h}_{c_{i}}^{H} \oplus_{H} \boldsymbol{h}_{c_{j}}^{H} = \frac{\big(1 + 2
    (\boldsymbol{h}_{c_{i}}^{H})^{T} \boldsymbol{h}_{c_{j}}^{H}
    + \norm{\boldsymbol{h}_{c_{j}}^{H}}_{2}^{2}\big)\boldsymbol{h}_{c_{i}}^{H} + (1 - \norm{\boldsymbol{h}_{c_{i}}^{H}}_{2}^{2})\boldsymbol{h}_{c_{j}}^{H}}
    {1 + 2
    (\boldsymbol{h}_{c_{i}}^{H})^{T}\boldsymbol{h}_{c_{j}}^{H}
    + \norm{\boldsymbol{h}_{c_{i}}^{H}}_{2}^{2}\norm{\boldsymbol{h}_{c_{j}}^{H}}_{2}^{2}} 
\end{equation}
where both the weight matrix $\boldsymbol{W}_{h_{r_{I\mathcal{O}k}}}$ and bias $\boldsymbol{b}_{h_{r_{I\mathcal{O}k}}}$ are specific to each relation $k$ in $R_{IO}$ and reserved for the ontology $\boldsymbol{h}_{r_{I\mathcal{O}k}}$.

\paragraph{\normalfont{\textbf{Bridge Node Loss Function.}}}
The bridge-node model uses a hinge loss function as a combination of the entity's ontology-specific loss, $\mathrm{ontoLoss}_{\boldsymbol{h}_{
r_{I\mathcal{O}k}}}$, and instance-specific loss, $\mathrm{instLoss}_{\boldsymbol{\theta}_{
r_{Ik}}}$, that is maximized for $R_{I} \cup R_{IO}$ which contains all triples associated with the bridge nodes.
Positive triples are denoted $\mathrm{pt}_{B, I} = (\boldsymbol{h}_{e_{i}}^{B}, \boldsymbol{\theta}_{
r_{Ik}}, \boldsymbol{h}_{e_{j}}^{B})$ and
$\mathrm{pt}_{B, IO} = (\boldsymbol{h}_{e_{i}}^{B}, \boldsymbol{h}_{
r_{IOk}}, \boldsymbol{h}_{c_{j}}^{B})$, and negative triples are denoted  $\mathrm{nt}_{B, I} = (\boldsymbol{h}_{e_{i}}^{B'}, \boldsymbol{\theta}_{
r_{Ik}}, \boldsymbol{h}_{e_{j}}^{B'})$ and
$\mathrm{nt}_{B, IO} = (\boldsymbol{h}_{e_{i}}^{B'}, \boldsymbol{h}_{
r_{IOk}}, \boldsymbol{h}_{c_{j}}^{B'})$ with loss function defined as follows:
\begin{multline}
    \mathrm{ontoLoss}_{\boldsymbol{h}_{
r_{I\mathcal{O}k}}}(\mathrm{pt_{B, I\mathcal{O}}}, \mathrm{nt_{B, I\mathcal{O}}}) \\ = \mathrm{max}\big(0, \gamma^{R_{I\mathcal{O}}} + \phi^{H}_{obs}(\mathrm{pt}_{B, I\mathcal{O}}) - \phi^{H}_{corr}(\mathrm{nt}_{B, I\mathcal{O}})\big)
\end{multline}
\begin{multline}
    \mathrm{instLoss}_{\boldsymbol{\theta}_{r_{Ik}}}(\mathrm{pt_{B, I}}, \mathrm{nt_{B, I}}) \\ = \mathrm{max}\big(0, \gamma^{R_{I}} + \phi^{S}_{\mathrm{obs}}(\mathrm{pt}_{B, I}) - \phi^{S}_{\mathrm{corr}}(\mathrm{nt}_{B, I})\big)
\end{multline}
\begin{multline}
    L_{\mathrm{bridge}}^{R_{I} \cup R_{IO}} = \frac{1}{|R_{I} \cup R_{IO}|} \sum_{(\mathrm{pt_{B, I}}, \mathrm{pt_{B, I\mathcal{O}}} \in R_{I} \cup R_{IO}) \land (\mathrm{nt_{B, I}}, \mathrm{nt_{B, I\mathcal{O}}} \notin R_{I} \cup R_{IO})}\big(\\
    {\mathrm{ontoLoss}_{\boldsymbol{h}_{r_{I\mathcal{O}k}}}(\mathrm{pt_{B, I\mathcal{O}}}, \mathrm{nt_{B, I\mathcal{O}}}) + \mathrm{instLoss}_{\boldsymbol{\theta}_{
r_{Ik}}}}(\mathrm{pt_{B, I}}, \mathrm{nt_{B, I}})\big)
\end{multline}
The combination loss function above enables bridge nodes to learn from both spaces of intersection of $\mathbb{S}^{d}$ and $\mathbb{H}^{d}$.

\subsection{Training}
\label{subsec:mgs-training}

This section details the training framework of \ourmodel, for representing two-view KGs, described in Section~\ref{subsec:dgs-modeling}. We describe, for each epoch, the training of each node in the two-view KG which includes (1) the forward propagation step, (2) the loss function, and (3) the backward propagation step to optimize parameters. Algorithm 1 provides a summary of the framework.

\paragraph{\normalfont{\textbf{Embedding Initialization.}}} We randomly initialize all embeddings of $e$ and $r$ in polar coordinates: $\boldsymbol{\theta}_{e_{i}} \in \mathbb{R}^{d-1} \leftarrow \mathrm{Unif}([0, 2\pi))^{d-1}$ and $\boldsymbol{\theta}_{r_{Ik}}, \boldsymbol{\theta}_{r_{\mathcal{O}k}} \in \mathbb{R}^{d-1} \leftarrow \mathrm{Unif}([0, 2\pi))^{d-1}$, then convert entity embeddings to their corresponding Cartesian representation: $\boldsymbol{h}_{e_{i}} = \mathrm{MPC}(\boldsymbol{\theta}_{e_{i}})$. Link $\boldsymbol{h}_{r_{I\mathcal{O}k}}$ is a randomly sampled position on the Poincare disk. Refer to Section~\ref{pc-framework} in Supplements for details about the conversion procedure, describing both polar-Cartesian coordinate conversion (MPC) and Cartesian-polar coordinate conversion (MCP). For the instance-view model, we also choose a value for norm $w^{S}$, that is sampled from a uniform distribution: $w^{S}: w^{S} \in [0, 1) \rightarrow \mathrm{Unif}([0,1))$ for all entities, and for the ontology-view model, we choose a value for norm $w_{\boldsymbol{h}_{c_{i}}}^{H}$, assigned uniformly at random per entity. We set the curvature values of the spherical and hyperbolic spaces as $K_{S} = 1$ and $K_{H} = -1$ respectively. We leave the non-trivial problem of learning optimal curvatures as future work.

\paragraph{\normalfont{\textbf{Training Procedure for Instance View KG}}}Parameter optimization is performed using Riemannian stochastic gradient descent (RSGD) for the spherical space as follows for entity embedding and relational embedding updates respectively. To ensure that the updated entity embedding remains in the norm-$w^{S}$ space, we perform a rescaling operation, $\mathrm{proj}_{S}$, to project out-of-boundary embeddings back to the surface of the $w^{S}$-ball.
\begin{equation}
\mathrm{proj}_{S}(\boldsymbol{z})=\begin{cases}
          w^{S} \cdot \frac{\boldsymbol{z}}{\norm{\boldsymbol{z}} } \quad &\text{if} \, \norm{\boldsymbol{z}} \neq w^{S} \\
          \boldsymbol{z} \quad &\text{otherwise} \\
     \end{cases}
\end{equation}
\begin{equation}
    r(\boldsymbol{h}_{e_{i,t}}^{S}, L_{\mathrm{inst}}^{R_{I}}) = \big(1 + \frac{\boldsymbol{h}_{e_{i,t}}^{S^{T}}\nabla{L_{\mathrm{inst}}^{R_{I}}}(\boldsymbol{h}_{e_{i,t}}^{S})}{\norm{\nabla{L_{\mathrm{inst}}^{R_{I}}}(\boldsymbol{h}_{e_{i,t}}^{S})}}\big)
    (I - \boldsymbol{h}_{e_{i,t}}^{S}\boldsymbol{h}_{e_{i,t}}^{S^{T}})
\end{equation}
\begin{equation}
    \boldsymbol{h}_{e_{i}, t+1}^{S} \leftarrow \mathrm{proj}_{S}\big(-\eta_{t} \cdot r(\boldsymbol{h}_{e_{i,t}}^{S}, L_{\mathrm{inst}}^{R_{I}})\nabla{L_{\mathrm{inst}}^{R_{I}}}(\boldsymbol{h}_{e_{i,t}}^{S})\big)
\end{equation}
\begin{equation}
    \boldsymbol{\theta}_{r_{Ik}, t+1}^{S} \leftarrow -\eta_{t} \cdot r(\boldsymbol{\theta}_{r_{Ik},t}^{S}, L_{\mathrm{inst}}^{R_{I}})\nabla{L_{\mathrm{inst}}^{R_{I}}}(\boldsymbol{\theta}_{r_{Ik},t}^{S})
\end{equation}

\paragraph{\normalfont{\textbf{Training procedure for concepts.}}} 

Parameter optimization is performed using RSGD for the hyperbolic space as follows for concept embedding and relational embedding updates respectively, where the corresponding concept norm space, $w^{H}_{c_{i}}$, is also learned through RSGD by updating embeddings of $\boldsymbol{h}_{c_{i}}^{H}$. Diagonal relational matrix $\boldsymbol{R}$ is also updated through RSGD, and we update scalar biases $b_{c_{i},t+1}^{H},b_{c_{j},t+1}^{H}$ through stochastic gradient descent. 
\begin{equation}
    \boldsymbol{h}_{c_{i, t+1}}^{H} \leftarrow \boldsymbol{h}_{c_{i, t}}^{H} - \eta_{t}(\frac{1 - \norm{\boldsymbol{h}_{c_{i, t}}^{H}}^{2}}{2})^{2}\nabla{L_{\mathrm{onto}}^{R_{\mathcal{O}} \cup R_{I\mathcal{O}}}}(\boldsymbol{h}_{c_{i,t}}^{H})
\end{equation}
\begin{equation}
    \boldsymbol{h}_{r_{\mathcal{O}k}, t+1}^{H} \leftarrow \boldsymbol{h}_{r_{\mathcal{O}k}, t}^{H} - \eta_{t}(\frac{1 - \norm{\boldsymbol{h}_{r_{\mathcal{O}k}, t}^{H}}^{2}}{2})^{2}\nabla{L_{\mathrm{onto}}^{R_{\mathcal{O}} \cup R_{I\mathcal{O}}}}(\boldsymbol{h}_{r_{\mathcal{O}k},t}^{H})
\end{equation}
\begin{equation}
    \boldsymbol{R}_{t+1} \leftarrow \boldsymbol{R}_{t} - \eta_{t}(\frac{1 - \norm{\boldsymbol{R}_{t}}^{2}}{2})^{2}\nabla{L_{\mathrm{onto}}^{R_{\mathcal{O}} \cup R_{I\mathcal{O}}}}(\boldsymbol{R}_{t})
\end{equation}
\begin{equation}
    b_{c_{i},t+1}^{H} \leftarrow b_{c_{i},t}^{H} - \eta_{t}\nabla{L_{\mathrm{onto}}^{R_{\mathcal{O}} \cup R_{I\mathcal{O}}}}(b_{c_{i},t}^{H})
\end{equation}
\begin{equation}
    b_{c_{j},t+1}^{H} \leftarrow b_{c_{j},t}^{H} - \eta_{t}\nabla{L_{\mathrm{onto}}^{R_{\mathcal{O}} \cup R_{I\mathcal{O}}}}(b_{c_{j},t}^{H})
\end{equation}

After the epoch's update of concept embeddings, we once again reset the value of the last dimension $d$ to 0 to satisfy the original framework constraint of the Poincaré disk. We also enforce that the angular dimensions of relational embeddings are in $[0, 2\pi)$.

\paragraph{\normalfont{\textbf{Training Procedure for Bridge Nodes}}}
Parameter optimization is performed using RSGD for the bridge space as follows for bridge entity embedding and relational embedding updates respectively. Ontology optimization, $\mathrm{ontoOpt}(\cdot)$, and instance optimization, $\mathrm{instOpt}(\cdot)$, are performed alternatively in batches for each of the two embedding types according to the type of link that the embedding is associated with, e.g., $R_{I}$ or $R_{IO}$. This enables the representation of bridge nodes to be informed by both $\mathbb{S}^{d}$ and $\mathbb{H}^{d}$.
\begin{equation}
    \mathrm{ontoOpt}(\boldsymbol{z}) = \boldsymbol{z} - \eta_{t}(\frac{1 - \norm{\boldsymbol{z}}^{2}}{2})^{2}\nabla{L_{\mathrm{bridge}}^{R_{I} \cup R_{IO}}}(\boldsymbol{z})
\end{equation}
\begin{equation}
    \mathrm{instOpt}(\boldsymbol{z}) = -\eta_{t} \cdot r(\boldsymbol{z}, L_{\mathrm{bridge}}^{R_{I} \cup R_{IO}})\nabla{L_{\mathrm{bridge}}^{R_{I} \cup R_{IO}}}(\boldsymbol{z})
\end{equation}
\begin{equation}
    \boldsymbol{h}_{e_{i},t+1}^{B} \leftarrow \mathrm{proj}_{B}\big(\mathrm{ontoOpt}(\boldsymbol{h}_{e_{i},t}^{B})\big) ||  \mathrm{proj}_{B}\big(\mathrm{instOpt}(\boldsymbol{h}_{e_{i},t}^{B})\big)
\end{equation}
\begin{equation}
    \boldsymbol{h}_{r_{I\mathcal{O}k},t+1} \leftarrow \mathrm{ontoOpt}(\boldsymbol{h}_{r_{I\mathcal{O}k},t})
\end{equation}
\begin{equation}
    \boldsymbol{\theta}_{r_{Ik},t+1} \leftarrow   \mathrm{instOpt}(\boldsymbol{\theta}_{r_{Ik},t})
\end{equation}
After each epoch's update of the instance optimization for $\boldsymbol{h}_{e_{i}}^{B}$, the value of the last dimension $d$ is reset to 0 and rescaled with $\mathrm{proj}_{B}(\boldsymbol{z})$ defined to be the same as $\mathrm{proj}_{S}(\boldsymbol{z})$ to ensure the intersection space constraint of the bridge entity model. Angular dimensions are also enforced to be in [0, 2$\pi$).

\begin{algorithm}[ht]
\small
	\SetKwInOut{Input}{\textbf{Input}}\SetKwInOut{Output}{\textbf{Output}}
	\Input{
	    set of entities $e$; set of relations $r$\\
	    instance-view entity to entity triples with links, $\boldsymbol{\theta}_{r_{Ik}}^{S}$\\
	    ontology-view concept to concept triples with links, $\boldsymbol{\theta}_{r_{\mathcal{O}k}}^{H}$\\
	    bridge entity to concept triples with links, ontology $\boldsymbol{h}_{r_{I\mathcal{O}k}}$\\
	    bridge entity to entity triples with links, instance $\boldsymbol{\theta}_{r_{Ik}}$
	}
	\Output{
		Updated embeddings,
		$\boldsymbol{\theta}_{r_{Ik}, \mathrm{EP}}^{S}$, $\boldsymbol{\theta}_{r_{\mathcal{O}k, \mathrm{EP}}}^{H}$, ontology $\boldsymbol{h}_{r_{I\mathcal{O}k, \mathrm{EP}}}$, and instance $\boldsymbol{\theta}_{r_{Ik}, \mathrm{EP}}$
		at final epoch $\mathrm{EP}$
	}
	\BlankLine
	\For{\normalfont{epoch} $\in$ (1, 2, ..., EP)}{
	    \textbf{Step 1*:}\\ 
	    Sample links from $\boldsymbol{\theta}_{r_{Ik}}^{S}$\\
	    Perform spherical update of entities: Section~\ref{subsec:inst-modeling}\\
	    \textbf{Step 2*:}\\ 
	    Sample links from $\boldsymbol{\theta}_{r_{\mathcal{O}k}}^{H}$\\
	    Perform hyperbolic update of concepts: Section~\ref{subsec:onto-modeling}\\
	    \textbf{Step 3*:}\\ 
	    Sample links from ontology $\boldsymbol{h}_{r_{I\mathcal{O}k}}$ and instance $\boldsymbol{\theta}_{r_{Ik}}$\\
	    Alternatively perform spherical and hyperbolic  updates of bridge nodes: Section~\ref{subsec:model-bridge}
	}
	
	
	\Return{$\boldsymbol{\theta}_{r_{Ik}, \mathrm{EP}}^{S}$, $\boldsymbol{\theta}_{r_{\mathcal{O}k, \mathrm{EP}}}^{H}$, \normalfont{ontology} $\boldsymbol{h}_{r_{I\mathcal{O}k, \mathrm{EP}}}$, \normalfont{and instance} $\boldsymbol{\theta}_{r_{Ik}, \mathrm{EP}}$}
	\caption{Overall training procedure of \ourmodel. ``*'' indicates that the three steps can be performed in any order.}
\end{algorithm}

\vspace{-4mm}
\section{Experiments}
\label{sec:exp}
In this section, we evaluate \ourmodel on two KG tasks: the triple completion task on each of the instance and ontology views of the KG and the entity typing task to test quality of the learned bridge space in communicating between each view of the KG. We also provide a case study on entity typing for different variants of \ourmodel by embedding on other combinations of geometric spaces. 
Further, we provide a visualization of embeddings before and after the learning process projected onto the 3-D geometric space of \ourmodel.
\subsection{Datasets}
We utilize the datasets of  YAGO26K-906 and DB111K-174 since they have the two-view KG setting unlike other datasets for KG embeddings that consider solely an instance-view~\cite{FB15k-237} or ontology-view~\cite{WIN18RR}. YAGO26K-906 and DB111K-174 are prepared from~\cite{joie}, which are extracted from YAGO~\cite{yago} and
DBpedia~\cite{db} respectively. Refer to~\cite{joie} for the detailed construction process. Table~\ref{tab:dataset-stat} provides dataset statistics and Table~\ref{tab:data-split} provides data splits for both datasets on both KG tasks. It can be observed that the instance-view contains many more triples than the ontology-view and that DB111K-174 contains a larger proportion of entity-concept triples (10.35\%) compared to YAGO26K-906 (2.43\%).

\begin{table}
\centering
\caption{\textmd{\small{
  Dataset statistics for entities $E$, concepts $C$ and their relations. $E-E$ denotes entity-entity links, $C-C$ denotes concept-concept links, and $E$-$C$ denotes entity-concept links.}}}
 \vspace{-2mm}
{
\footnotesize
\begin{tabular}{c|cc|ccc}
\bottomrule
Dataset &
\multicolumn{2}{c|}{Nodes} &
\multicolumn{3}{c}{Relations} \\
  & \#$E$ & \#$C$ & \#$E$-$E$: $R_{I}$ & \#$C$-$C$: $R_{O}$ & \#$E$-$C$: $R_{IO}$\\
\hline
 YAGO26K-906 & 26,078 & 906 &  390,738 &  8,962 & 9,962 \\
 DB111K-174 & 111,762 & 174 & 863,643 &  763 & 99,748 \\
 \toprule
\end{tabular}
\vspace{-2mm}
}
\label{tab:dataset-stat}
\end{table}

\vspace{-2mm}
\begin{table}
\centering
\caption{\textmd{\small{
  Data splits for triple completion and entity typing. We provide splits for all KG triples in $R_{I}, R_{O}, R_{IO}$ for train(tr), validation(v), and test(ts).}}}
\vspace{-2mm}
{
\footnotesize
\begin{tabular}{c|ccc}
\bottomrule
 & &
 YAGO26K-906 &  \\
Task &
Tr($R_{I}/R_{O}/R_{IO}$) & V($R_{I}/R_{O}/R_{IO}$) & Ts($R_{I}/R_{O}/R_{IO}$) \\
\hline
Triple Completion & 332,128/7,618/8,691 & 19,536/448/485 & 39,074/896/1,019\\
Entity Typing & 211,346/4,876/5,379 & 23,549/543/598 & 156,311/3,592/3,985\\
\hline
\hline
 & &
 DB111K-174 &  \\
Task &
Tr($R_{I}/R_{O}/R_{IO}$) & V($R_{I}/R_{O}/R_{IO}$) & Ts($R_{I}/R_{O}/R_{IO}$) \\
\hline
Triple Completion & 734,096/648/84,864 & 43,182/38/5,018 & 86,365/77/10,131 \\
Entity Typing & 466,538/462/53,863 & 51,828/46/5,985 & 345,504/337/39,900 \\
\toprule
\end{tabular}
}
\label{tab:data-split}
\vspace{-1mm}
\end{table}

\subsection{Models}
\subsubsection{Baselines} We compare \ourmodel to state-of-the-art neural network embedding models, which include Euclidean, non-Euclidean, and product space KGE models, as well as GNN-based models for KG completion and entity typing.
\begin{itemize}
    \item \transe~\cite{transe}, one of the first KGE models, which simply captures the relationship between entities as a translation.
    
    \item \distmult~\cite{distmult}, a matrix factorization KGE model, modeling the relationship between entities via multiplication.
    
    \item \complex~\cite{complex}, a KGE model that extends \distmult into the complex number field.
    
    \item \rotate~\cite{rotatE}, a recent KGE model, based on the rotation assumption where a relation is a rotation from the subject to the object in the complex vector space.
    
    \item \joie~\cite{joie} and \mtwognn~\cite{m2gnn}: Refer to Section~\ref{gen-kg-mod} where this is discussed. 
    
    \item \hyperkg~\cite{hyperkg}, a KGE model extending translational KGE methods to the Poincaré-ball model of hyperbolic geometry.
    
    \item \hake~\cite{hake}, which extends \rotate by having relations combining modulus scaling with rotation.
    
    \item \cone~\cite{cone}, a KGE model embedding entities into hyperbolic cones and relations as transformations between cones.
    
    \item \textbf{\rehf/\roth/\atth}~\cite{atth}, which are hyperbolic KGE models that combine hyperbolic spaces using hyperbolic attention, where \rehf and \roth are variants of \atth using only reflections and rotations respectively.
    
    \item \hgcn~\cite{hgcn}, a hyperbolic GCN model utilizing Riemannian geometry and the hyperboloid model.
    
    \item \hyperka~\cite{hyperka}, which extends GCNs from the Euclidean to hyperbolic space using the Poincaré ball model.
    
\end{itemize}

\subsubsection{\ourmodel Variants} We describe variant models of \ourmodel below. 

\begin{itemize}
    \item \textbf{\ourmodel-RO-FC}, which is \ourmodel with the Riemannian operator (RO) used for vector transformation instead of our proposed closed spherical space vector transformation procedure in Section~\ref{subsec:inst-modeling}, and with fixed center (FC) of spherical ball at 0, which is the same center as the Poincaré disk. For single geometric spaces of $\mathbb{S}^{d}$ and $\mathbb{H}^{d}$ for non-bridge nodes and concepts, the Riemannian operator is performed as a retraction operation to the tangent Euclidean spaces. However, we extend the Riemannian operator when performing retraction for the intersection ring of the bridge space, which is formed by intersecting the spherical ball's surface and Poincaré disk. This is described in the Supplements.   
    
    \item \textbf{\ourmodel-SO-FC}, which is \ourmodel with the proposed closed spherical space operator (SO), and with FC of spherical ball at 0, which is the same center as the Poincaré disk.
    
    \item \textbf{\ourmodel (ours)}, which is \ourmodel with the proposed closed SO, and with learnable center (LC) of spherical ball, which for simplicity of constructing the intersection space for bridge nodes, is set to the last dimension to only allow for vertical shift. Note that we do not need to also make the center of the Poincaré disk learnable as this shift is already introduced with making one of the centers learnable. For learning the spherical center, $\omega$, in the model, we follow the same training procedure for Section~\ref{subsec:mgs-training} but for the non-bridge entities and bridge entities, we perform the operations by temporarily shifting to the center 0 (e.g., $-\omega$ shift), then shift back to the new center $\omega$ (e.g., $+\omega$ shift) after the updates are performed.  
    
\end{itemize}

\vspace{-1mm}
\subsubsection{\ourmodel Ablation Models} We study different ablations models of \ourmodel, which are formed by utilizing different combinations of manifold spaces for each type of link of \ourmodel($R_{I}/R_{O}$) in the two-view KG including the spherical space, $\mathbb{S}^{d}$, hyperbolic space using the Poincaré model, $\mathbb{H}^{d}$, or Euclidean space, $\mathbb{E}^{d}$. These include: \textbf{\ourmodel ($\mathbb{S}^{d}/\mathbb{S}^{d}$)}, \textbf{MGS ($\mathbb{H}^{d}/\mathbb{H}^{d}$)}, \textbf{\ourmodel ($\mathbb{E}^{d}/\mathbb{H}^{d}$)}, and \textbf{\ourmodel ($\mathbb{S}^{d}/\mathbb{E}^{d}$)}. Since $R_{IO}$ is always at the intersection of the two spaces $R_{I}$ and $R_{O}$, we do not need to specify the geometric space separately. 

\vspace{-2mm}
\subsection{Evaluation}
\begin{table*}[!htb]
\caption{\textmd{\small{
  Results of KG triple completion. $\boldsymbol{M}$ denotes the $d$-dimensional manifold space, $R_{I}$ are entity links, $R_{O}$ are concept links. For each group of models, the best results are bold-faced. The overall best results on each dataset are underscored.}}}
 \vspace{-2mm}
\centering
{
\footnotesize
\begin{tabular}{c|c|c|ccc|ccc||ccc|ccc}
\bottomrule
 & & {Datasets} &
\multicolumn{6}{c||}{YAGO26K-906} &
\multicolumn{6}{c}{DB111K-174}
\\
Type & $\boldsymbol{M}$ & {Graphs} &
\multicolumn{3}{c}{$R_I$ KG Completion} &
\multicolumn{3}{c||}{$R_O$ KG Completion} & 
\multicolumn{3}{c}{$R_I$ KG Completion} &
\multicolumn{3}{c}{$R_O$ KG Completion}
\\
& & {Metrics} & 
MRR & H@1 & H@10 &
MRR & H@1 & H@10 &
MRR & H@1 & H@10 &
MRR & H@1 & H@10 
\\
\hline
\hline
& $\mathbb{E}^{d}$ & {\transe} & 0.187 & 13.73 & 35.05 & 0.189 & {14.72} & 24.36 & 0.318 & 22.70 & 48.12 & 0.539 & 47.90 & 61.84 \\
& $\mathbb{E}^{d}$ & {\distmult} & 0.288 & 24.06 & 31.24 & 0.156 & 14.32 & 16.54 & 0.280 & 27.24 & 29.70 & 0.501 & 45.52 & 64.73\\
 non-GNN based& $\mathbb{C}^{d}$ & \complex & 0.291 & 24.85 & 37.28 & 0.180 & 14.83 & 22.97 & 0.321 & 27.39 & 46.63 & 0.549 & 47.80 & 62.23 \\
KGE models & $\mathbb{C}^{d}$ & {\rotate} & 0.302 & \textbf{25.31} & 42.17 & 0.228 & 16.35 & 27.28 & 0.356 & 29.31 & 54.60 & 0.557 & 49.16 & 68.19 \\
& $\mathbb{E}^{d}$ & {\joie} & \textbf{0.316} & 24.62 & \textbf{51.85} & 0.289 & 18.66 & 39.13 & \textbf{0.479} & \textbf{35.21} & \textbf{72.38} & 0.602 & 52.48 & 79.71 \\
& $\mathbb{H}^{d}$ & {\hyperkg} & 0.215 & 18.35 & 36.02 & 0.174 & 14.50 & 23.26 & 0.302 & 23.31 & 46.72 & 0.542 & 47.59 & 62.11 \\
& $\mathbb{H}^{d}$ & {\hake} & 0.293 & 23.04 & 40.19 & 0.301 & 19.27 & 41.09 & 0.391 & 31.10 & 60.46 & 0.638 & 55.69 & 81.07 \\
& $\mathbb{H}^{d}$ & {\cone} & 0.299 & 23.56 & 41.23 & \textbf{0.313} & \textbf{20.05} & \textbf{41.80} & 0.422 & 33.69 & 68.12 & \textbf{0.639} & \textbf{55.89} & \textbf{81.45} \\
\hline
& $\mathbb{H}^{d}$ & {\rehf} & 0.282 & 23.19 & 40.52 & 0.298 & 19.70 & 41.26 & 0.407 & 30.06 & 66.93 & 0.622 & 55.35 & 81.09 \\
& $\mathbb{H}^{d}$ & {\roth} & 0.295 & 23.50 & 41.03 & 0.308 & 19.97 & \textbf{41.78} & 0.418 & 30.18 & 67.05 & 0.639 & 55.82 & 81.44 \\
GNN-based models& $\mathbb{H}^{d}$ & {\atth} & 0.298 & 23.43 & 41.20 & 0.310 & 19.99 & 41.53 & 0.419 & 30.10 & 66.58 & 0.629 & 55.37 & 81.39 \\
& $\mathbb{H}^{d}$ & {\hgcn} & 0.307 & 23.04 & 40.25 & 0.302 & 19.38 & 40.49 & 0.396 & 31.54 & 61.78 & 0.638 & 55.81 & 81.60 \\
& $\mathbb{H}^{d}$ & {\hyperka} & 0.320 & 26.71 & 52.09 & 0.305 & 18.83 & 40.28 & 0.486 & 35.79 & 72.33 & 0.613 & 53.36 & 80.59 \\
& $\mathbb{P}^{d}$ & {\mtwognn} & \textbf{0.347} & \textbf{29.63} & \textbf{54.28} & \textbf{0.341} & \textbf{23.70} & 42.19 & \textbf{0.506} & \textbf{36.52} & \textbf{73.11} & \textbf{0.644} & \textbf{56.82} & \textbf{83.01} \\
\hline
\rowcolor{shadecolor} & -- & \ourmodel ($\mathbb{S}^{d}/\mathbb{S}^{d}$) & \textbf{0.338} & \textbf{27.15} & \textbf{53.20} & 0.318 & 20.36 & 41.02 & 0.491 & 34.58 & 71.40 & 0.606 & 53.29 & 80.17\\
\rowcolor{shadecolor} Ablation variants: & -- & \ourmodel ($\mathbb{H}^{d}/\mathbb{H}^{d}$) & 0.314 & 25.11 & 52.02 & \textbf{0.358} & \textbf{24.61} & \textbf{43.28} & \textbf{0.502} & \textbf{35.79} & \textbf{73.61} & \textbf{0.663} & \textbf{57.59} & \textbf{84.16}\\
\rowcolor{shadecolor} \ourmodel ($R_{I}/R_{O}$) & -- & \ourmodel ($\mathbb{E}^{d}/\mathbb{H}^{d}$) & 0.327 & 25.32 & 52.89 & 0.343 & 23.95 & 41.62 & 0.498 & 35.11 & 72.37 & 0.640 & 56.17 & 82.68\\
\rowcolor{shadecolor} & -- & \ourmodel ($\mathbb{S}^{d}/\mathbb{E}^{d}$) & 0.322 & 24.91 & 52.36 & 0.297 & 19.43 & 40.61 & 0.484 & 33.29 & 73.54 & 0.619 & 53.72 & 80.51\\
\hline
\rowcolor{shadecolor} & -- & {\ourmodel-RO-FC} & 0.352 & 29.79 & 55.21 & 0.364 & 25.04 & 43.27 & 0.518 & 37.65 & 73.97 & 0.681 & 59.23 & 84.16\\
\rowcolor{shadecolor} \ourmodel variants & -- & {\ourmodel-SO-FC} & 0.364 & \underline{\textbf{30.15}} & 55.93 & 0.369 & 25.81 & 44.18 & \underline{\textbf{0.536}} & 38.29 & 74.28 & 0.687 & 59.26 & \underline{\textbf{84.82}}\\
\rowcolor{shadecolor} & -- & \ourmodel (ours) & \underline{\textbf{0.366}} & \underline{\textbf{30.15}} & \underline{\textbf{56.06}} & \underline{\textbf{0.372}} & \underline{\textbf{25.88}} & \underline{\textbf{44.38}} & \underline{\textbf{0.536}} & \underline{\textbf{38.31}} & \underline{\textbf{74.85}} & \underline{\textbf{0.690}} & \underline{\textbf{59.88}} & \underline{\textbf{84.82}}\\
\toprule
\end{tabular}
\vspace{-3mm}
}
\label{tab:link}

\end{table*}

In this section, we detail our evaluation on the tasks of KG triple completion and entity typing. The goal of triple completion is to construct the missing relation facts in a KG structure. Specifically, we test constructing a missing target node, from each of the ontology or instance views: or queries $(e_{i}, r_{k}, ?e_{j})$ and $(c_{i}, r_{k}, ?c_{j})$, such that each model evaluated is trained on the entire two-view KG. The goal of entity typing is to predict the concepts that correspond to the entities, or queries $(e_{i}, r_{k}, ?c_{j})$. 

Using plausibility scores to rank each test candidate set, for each task, we report results for the evaluation metrics of mean reciprocal rank ($MRR$) and Hits@, e.g., $Hits@1$, $Hits@3$, $Hits@10$. Table~\ref{tab:data-split} reports our data splits for each task. For triple completion, this is chosen to be embedding distance of the source node to the missing target node modeled under the relation, and for entity typing, this is chosen to be the embedding distance from the entity's representation in the concept space to the concept. \cite{joie} provides more details on the evaluation procedure. For evaluation consistency, for both tasks, model training hyperparameters are chosen for dimensionality $d \in$ \{50, 100, 200, 300\} for all triples, learning rate $\eta \in$ \{5e-4, 1e-3, 1e-2, 1e-1\} and margins $\gamma \in$ \{0.5, 1\}. Further, different batch sizes and epochs are used according to the type and size of the graphs.

\vspace{-1mm}
\paragraph{\normalfont{\textbf{KG Triple Completion Results.}}} Results are reported in Table~\ref{tab:link}. \ourmodel outperforms all of the baseline models on both datasets. \ourmodel achieves an average performance gain over all baselines by 32.36\% on $MRR$, 27.59\% on $Hit@1$, and 29.17\% on $Hit@10$ for the instance-view completion across both YAGO26K-906 and DB111K-174. \ourmodel achieves an average performance gain over all baselines by 23.43\% on $MRR$, 28.41\% on $Hit@1$, and 18.11\% on $Hit@10$ for the ontology-view completion across both YAGO26K-906 and DB111K-174.

It can be observed that in both the instance and ontology views on both datasets, the hyperbolic-based KGE models outperform their Euclidean and complex space counterparts. Further, hyperbolic KGE models perform better on ontology view than instance view likely due to there being prevalence of hierarchy in the ontology. It is also seen that using multiple geometric spaces is more effective than using a single geometric space. For GNN-based models, \mtwognn, which uses a product space, $\mathbb{P}^{d}$ combining Euclidean, spherical, and hyperbolic spaces, outperforms the models using only one of the spaces. Compared to \mtwognn, the most competitive baseline model in most cases, \ourmodel shows significant improvement of 3.25\% on $MRR$, 5.15\% on $Hit@1$, and 2.73\% on $Hit@10$ on average for both YAGO26K-906 and DB111K-174. This is likely because hierarchy in the KG is better modeled through the hyperbolic space than with influence from the spherical or Euclidean space, and cyclic links are better modeled through the space than with influence from the hyperbolic or Euclidean space. For bridge entities involved in both hierarchical and cyclic links, we see it is beneficial to model them in an intersection space to better model both of these properties.       

\begin{table}
\caption{\textmd{\small{
  Results of entity typing. For each group of models, the best results are bold-faced. The overall best results on each dataset are underscored.}}}
\vspace{-2mm}
\footnotesize
\setlength\tabcolsep{2.5pt}
{
\begin{tabular}{c|ccc|ccc}
\bottomrule
{Datasets} &
\multicolumn{3}{c|}{YAGO26K-906} &
\multicolumn{3}{c}{DB111K-174}
\\
{Metrics} & 
MRR & Acc. & Hit@3 &
MRR & Acc. & Hit@3
\\
\hline
\hline
{\transe} 	& 0.144 & 7.32 	& 35.26	& 0.503 & 43.67 & 60.78\\
{\distmult} 	& 0.411 & 36.07 & 55.32 & 0.551 & 49.83 & 68.01\\ 
{\complex} 	& 0.519 & 43.08 & 59.28 & 0.583 & 55.07 & 70.17\\ 
{\rotate} 	& 0.673 & 58.24 & 73.95 & 0.721 & 61.48 & 75.67\\
{\joie} 	& 0.899 & 85.72 & 96.02 & 0.859 & 75.58 & 96.02\\ 
{\hyperkg} 	& 0.627 & 55.39 & 65.64 & 0.736 & 61.20 & 74.29\\ 
{\hake} 	& 0.905 & 87.42 & 96.37 & 0.866 & 76.04 & 96.22\\ 
{\cone} 	& \textbf{0.912} & \textbf{87.51} & \textbf{96.39} & \textbf{0.869} & \textbf{76.85} & \textbf{96.38}\\ 
\hline
{\rehf} 	& 0.907 & 87.49 & 96.37 & 0.867 & 76.26 & 96.31\\ 
{\roth} 	& 0.909 & 87.49 & 96.39 & 0.868 & 76.55 & 96.36\\ 
{\atth} 	& 0.910 & 87.50 & 96.38 & 0.868 & 76.50 & 96.33\\ 
{\hgcn} 	& 0.905 & 87.44 & 96.37 & 0.867 & 76.11 & 96.27\\ 
{\hyperka} 	& 0.918 & 87.76 & 96.45 & 0.871 & 76.65 & 96.50\\ 
{\mtwognn} 	& \textbf{0.922} & \textbf{88.16} & \textbf{97.01} & \textbf{0.880} & \textbf{77.58} & \textbf{96.94}\\ 
\hline
\rowcolor{shadecolor} {\ourmodel ($\mathbb{S}^{d}/\mathbb{S}^{d}$)} 	& 0.919 & 87.94 & 96.81 & 0.875 & 77.03 & 96.78\\ 
\rowcolor{shadecolor} {\ourmodel ($\mathbb{H}^{d}/\mathbb{H}^{d}$)} 	& \textbf{0.924} & \textbf{88.12} & \textbf{97.01} & \textbf{0.887} & \textbf{77.53} & \textbf{96.95}\\ 
\rowcolor{shadecolor} {\ourmodel ($\mathbb{E}^{d}/\mathbb{H}^{d}$)} 	& 0.922 & 88.07 & 96.91 & 0.883 & 77.37 & 96.58\\ 
\rowcolor{shadecolor} {\ourmodel ($\mathbb{S}^{d}/\mathbb{E}^{d}$)} 	& 0.917 & 87.68 & 96.41 & 0.866 & 76.33 & 96.48\\ 
\hline
\rowcolor{shadecolor} {\ourmodel-RO-FC} 	& 0.930 & 88.47 & 97.32 & 0.891 & 77.81 & 97.04\\ 
\rowcolor{shadecolor} {\ourmodel-SO-FC} 	& 0.938 & 89.02 & 98.11 & \underline{\textbf{0.895}} & 77.92 & 97.40\\ 
\rowcolor{shadecolor} {\ourmodel (ours)} 	& \underline{\textbf{0.939}} & \underline{\textbf{89.07}} & \underline{\textbf{98.18}} & \underline{\textbf{0.895}} & \underline{\textbf{77.94}} & \underline{\textbf{97.47}}\\ 
\toprule
\end{tabular}
}
\label{tab:type}
\end{table}

\vspace{-1mm}
\paragraph{\normalfont{\textbf{Entity Typing Results.}}}
Results are reported in Table~\ref{tab:type}. Consistent with the KG completion results, it can be seen that \ourmodel variants outperform the baseline models on both datasets for entity typing. \ourmodel achieves an average performance gain over all baselines of 24.49\% on $MRR$, 26.39\% on $Acc$, and 18.78\% on $Hit@3$ for YAGO26K-906. \ourmodel achieves an average performance gain over all baselines of 14.86\% on $MRR$, 13.74\% on $Acc$, and 12.15\% on $Hit@3$ for DB111K-174. Compared to the most competitive GNN-based model, \mtwognn, \ourmodel has nearly a 2\% gain in $MRR$ for each YAGO26K-906 and DB111K-174.

\vspace{-1mm}
\paragraph{\normalfont{\textbf{Ablation Studies.}}}
Tables~\ref{tab:link} and~\ref{tab:type} report our results for ablation studies in models belonging to the category \textit{Ablation variants}. It can be seen that using the spherical space for $R_{I}$ links learns a better representation than its hyperbolic or Euclidean space counterparts. This indicates the prevalence of cyclic relations between entities. Utilizing the hyperbolic space for $R_{O}$ and $R_{IO}$ links learns a better representation than its spherical or Euclidean space counterparts. This indicates the prevalence of hierarchical relations between concepts and entities-concepts. Interestingly, a fully hyperbolic-space model in general achieves better performance than its fully spherical counterpart, indicating that  YAGO26K-906 and DB111K-174 may contain relatively more hierarchical links than cyclic links. 
Further, learning a better representation in one view benefits the other view more if there are more $R_{IO}$ bridge links. For example, since DB111K-174 contains significantly more $R_{IO}$ links than YAGO26K-906 shown in Table~\ref{tab:dataset-stat}, it can be seen that \ourmodel ($\mathbb{H}^{d}/\mathbb{H}^{d}/\mathbb{H}^{d}$) has better performance against baselines on the instance view of DB111K-174 compared to the instance view of YAGO26K-906.

\vspace{-1mm}

\paragraph{\normalfont{\textbf{Visualizations.}}}

\begin{figure}[h]
    \vspace{-3mm}
    \centering
    \includegraphics[width=0.9\linewidth]{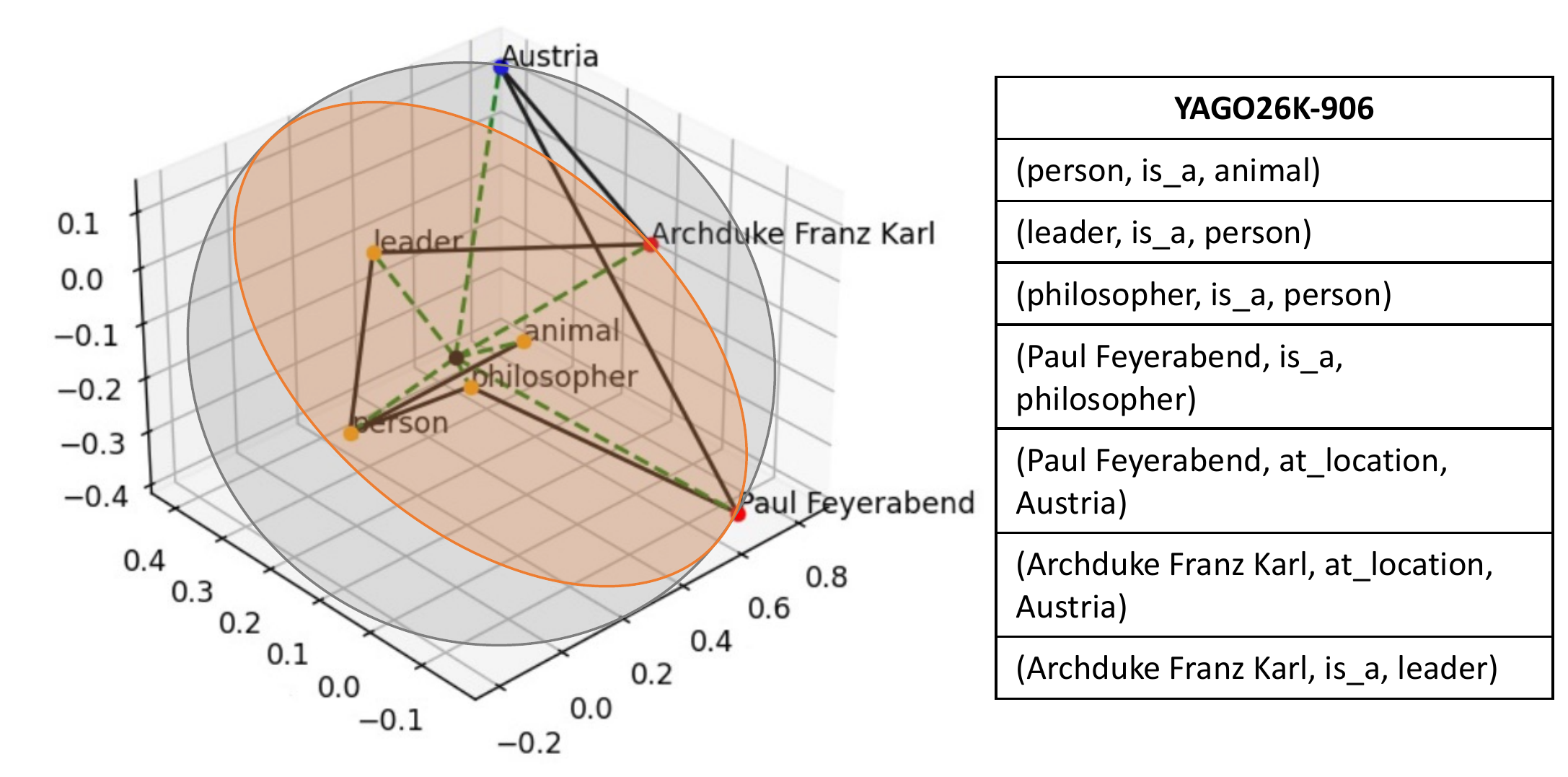}
    \caption{\textmd{Example of embeddings learned by \ourmodel. Solid lines connect entities and concepts as in the original dataset. Dashed lines connect every node to the origin to indicate closeness to origin.}
    }
    \label{fig:example-fig}
    \vspace{-4.5mm}
\end{figure}

We project the learned embeddings into 3-D space, and plot seven nodes with their relations in Figure~\ref{fig:example-fig}. Concepts that are higher in the ontological hierarchy such as \textit{animal} and \textit{person} are closer to the origin compared to lower-level concepts such as \textit{leader}, and entities tend to be farther away from the origin.

\vspace{-1mm}
\section{Conclusions}
We are among the first to explore utilization of different embedding spaces for different views of a knowledge graph. Entities in instance view often have cyclical connections and are hereby modeled via spherical embeddings; whereas concepts and bridge entities in ontology view tend to form a hierarchy and are hereby modeled in hyperbolic space. We propose the notion of bridge space to seamlessly model intersection of the two spaces in which bridge entities reside.  
In addition, we propose a set of closed spherical space operations to eliminate the need of projecting embeddings to the tangent Euclidean space. The resulting model, \ourmodel, is a unified framework that significantly outperforms all baselines, showing the effectiveness of capturing different structures in the knowledge graph using embedding spaces of different curvatures and properties.

Future directions include supporting geometric spaces of learnable curvature (to better capture the knowledge graph structure) and allowing a learnable offset between the origins of the two geometric spaces (to better accommodate uneven entity distribution and/or unbalanced ontology structures) 
We also aim to extend our dual-space model to a multi-space geometric embedding model. 

\vspace{-1mm}
\section{Acknowledgements}

This work was partially supported by NSF III-1705169, 1829071, 1937599, 2106859, 2119643; NIH R35-HL135772; NIBIB R01-EB027650; Amazon Research Awards; Cisco Research Grant USA000EP280889; NEC Gifts; Picsart Gifts; and Snapchat Gifts. The authors would also like to thank anonymous reviewers for their constructive feedback.


\vspace{-3mm}
\bibliographystyle{ACM-Reference-Format}
\bibliography{main}


\begin{thebibliography}{32}


\ifx \showCODEN    \undefined \def \showCODEN     #1{\unskip}     \fi
\ifx \showDOI      \undefined \def \showDOI       #1{#1}\fi
\ifx \showISBNx    \undefined \def \showISBNx     #1{\unskip}     \fi
\ifx \showISBNxiii \undefined \def \showISBNxiii  #1{\unskip}     \fi
\ifx \showISSN     \undefined \def \showISSN      #1{\unskip}     \fi
\ifx \showLCCN     \undefined \def \showLCCN      #1{\unskip}     \fi
\ifx \shownote     \undefined \def \shownote      #1{#1}          \fi
\ifx \showarticletitle \undefined \def \showarticletitle #1{#1}   \fi
\ifx \showURL      \undefined \def \showURL       {\relax}        \fi
\providecommand\bibfield[2]{#2}
\providecommand\bibinfo[2]{#2}
\providecommand\natexlab[1]{#1}
\providecommand\showeprint[2][]{arXiv:#2}

\bibitem[Bai et~al\mbox{.}(2021)]%
        {cone}
\bibfield{author}{\bibinfo{person}{Yushi Bai}, \bibinfo{person}{Rex Ying},
  \bibinfo{person}{Hongyu Ren}, {and} \bibinfo{person}{Jure Leskovec}.}
  \bibinfo{year}{2021}\natexlab{}.
\newblock \showarticletitle{Modeling Heterogeneous Hierarchies with
  Relation-specific Hyperbolic Cones}.
\newblock \bibinfo{journal}{\emph{NeurIPS}} (\bibinfo{year}{2021}).
\newblock


\bibitem[Balaževic´ et~al\mbox{.}(2019)]%
        {murp}
\bibfield{author}{\bibinfo{person}{Ivana Balaževic´}, \bibinfo{person}{Carl
  Allen}, {and} \bibinfo{person}{Timothy Hospedales}.}
  \bibinfo{year}{2019}\natexlab{}.
\newblock \showarticletitle{Multi-relational Poincaré Graph Embeddings}.
\newblock \bibinfo{journal}{\emph{NeurIPS}} (\bibinfo{year}{2019}).
\newblock


\bibitem[Beltrami(1897)]%
        {lorentz-orig}
\bibfield{author}{\bibinfo{person}{Eugenio Beltrami}.}
  \bibinfo{year}{1897}\natexlab{}.
\newblock \bibinfo{booktitle}{\emph{Teoria fondamentale degli spazii di
  curvatura costante.}}
\newblock \bibinfo{publisher}{Annali di Matematica Pura ed Applicata}.
\newblock


\bibitem[Berant and Liang(2014)]%
        {kg-sem-parsing}
\bibfield{author}{\bibinfo{person}{Jonathan Berant} {and}
  \bibinfo{person}{Percy Liang}.} \bibinfo{year}{2014}\natexlab{}.
\newblock \showarticletitle{Semantic Parsing via Paraphrasing}.
\newblock \bibinfo{journal}{\emph{ACL}} (\bibinfo{year}{2014}).
\newblock


\bibitem[Bonnabel(2013)]%
        {rsgd}
\bibfield{author}{\bibinfo{person}{Silvère Bonnabel}.}
  \bibinfo{year}{2013}\natexlab{}.
\newblock \showarticletitle{Stochastic gradient descent on Riemannian
  manifolds}.
\newblock \bibinfo{journal}{\emph{IEEE}} (\bibinfo{year}{2013}).
\newblock


\bibitem[Bordes et~al\mbox{.}(2013)]%
        {transe}
\bibfield{author}{\bibinfo{person}{Antoine Bordes}, \bibinfo{person}{Nicolas
  Usunier}, \bibinfo{person}{Alberto Garcia-Duran}, \bibinfo{person}{Jason
  Weston}, {and} \bibinfo{person}{Oksana Yakhnenko}.}
  \bibinfo{year}{2013}\natexlab{}.
\newblock \showarticletitle{Translating Embeddings for Modeling
  Multi-relational Data}.
\newblock \bibinfo{journal}{\emph{NeurIPS}} (\bibinfo{year}{2013}).
\newblock


\bibitem[Cannon et~al\mbox{.}(1997)]%
        {poincare-orig}
\bibfield{author}{\bibinfo{person}{James~W. Cannon},
  \bibinfo{person}{William~J. Floyd}, \bibinfo{person}{Richard Kenyon}, {and}
  \bibinfo{person}{Walter~R. Parry}.} \bibinfo{year}{1997}\natexlab{}.
\newblock \bibinfo{booktitle}{\emph{Hyperbolic Geometry}}.
\newblock \bibinfo{publisher}{MSRI}.
\newblock


\bibitem[Chami et~al\mbox{.}(2020)]%
        {atth}
\bibfield{author}{\bibinfo{person}{Ines Chami}, \bibinfo{person}{Adva Wolf},
  \bibinfo{person}{Da-Cheng Juan}, \bibinfo{person}{Frederic Sala},
  \bibinfo{person}{Sujith Ravi}, {and} \bibinfo{person}{Christopher Re´}.}
  \bibinfo{year}{2020}\natexlab{}.
\newblock \showarticletitle{Low-Dimensional Hyperbolic Knowledge Graph
  Embeddings}.
\newblock \bibinfo{journal}{\emph{ACL}} (\bibinfo{year}{2020}).
\newblock


\bibitem[Chami et~al\mbox{.}(2019)]%
        {hgcn}
\bibfield{author}{\bibinfo{person}{Ines Chami}, \bibinfo{person}{Rex Ying},
  \bibinfo{person}{Christopher Ré}, {and} \bibinfo{person}{Jure Leskovec}.}
  \bibinfo{year}{2019}\natexlab{}.
\newblock \showarticletitle{Hyperbolic Graph Convolutional Neural Networks}.
\newblock \bibinfo{journal}{\emph{NeurIPS}} (\bibinfo{year}{2019}).
\newblock


\bibitem[Dettmers et~al\mbox{.}(2018)]%
        {WIN18RR}
\bibfield{author}{\bibinfo{person}{Tim Dettmers}, \bibinfo{person}{Pasquale
  Minervini}, \bibinfo{person}{Pontus Stenetorp}, {and}
  \bibinfo{person}{Sebastian Riedel}.} \bibinfo{year}{2018}\natexlab{}.
\newblock \showarticletitle{Convolutional 2D knowledge graph embeddings}.
\newblock \bibinfo{journal}{\emph{AAAI}}.
\newblock


\bibitem[Gu et~al\mbox{.}(2018)]%
        {gu2018learning}
\bibfield{author}{\bibinfo{person}{Albert Gu}, \bibinfo{person}{Frederic Sala},
  \bibinfo{person}{Beliz Gunel}, {and} \bibinfo{person}{Christopher R{\'e}}.}
  \bibinfo{year}{2018}\natexlab{}.
\newblock \showarticletitle{Learning mixed-curvature representations in product
  spaces}. In \bibinfo{booktitle}{\emph{ICLR}}.
\newblock


\bibitem[Hao et~al\mbox{.}(2019)]%
        {joie}
\bibfield{author}{\bibinfo{person}{Junheng Hao}, \bibinfo{person}{Muhao Chen},
  \bibinfo{person}{Wenchao Yu}, \bibinfo{person}{Yizhou Sun}, {and}
  \bibinfo{person}{Wei Wang}.} \bibinfo{year}{2019}\natexlab{}.
\newblock \showarticletitle{Universal Representation Learning of Knowledge
  Bases by Jointly Embedding Instances and Ontological Concepts}.
\newblock \bibinfo{journal}{\emph{KDD}} (\bibinfo{year}{2019}).
\newblock


\bibitem[Huang et~al\mbox{.}(2019a)]%
        {kg-qa}
\bibfield{author}{\bibinfo{person}{Xiao Huang}, \bibinfo{person}{Jingyuan
  Zhang}, \bibinfo{person}{Dingcheng Li}, {and} \bibinfo{person}{Ping Li}.}
  \bibinfo{year}{2019}\natexlab{a}.
\newblock \showarticletitle{Knowledge Graph Embedding Based Question
  Answering}.
\newblock \bibinfo{journal}{\emph{WSDM}} (\bibinfo{year}{2019}).
\newblock


\bibitem[Huang et~al\mbox{.}(2019b)]%
        {general-kg}
\bibfield{author}{\bibinfo{person}{Xiao Huang}, \bibinfo{person}{Jingyuan
  Zhang}, \bibinfo{person}{Dingcheng Li}, {and} \bibinfo{person}{Ping Li}.}
  \bibinfo{year}{2019}\natexlab{b}.
\newblock \showarticletitle{Knowledge Graph Embedding Based Question
  Answering}.
\newblock \bibinfo{journal}{\emph{WSDM}} (\bibinfo{year}{2019}).
\newblock


\bibitem[Iyer et~al\mbox{.}(2022)]%
        {qa-gnn}
\bibfield{author}{\bibinfo{person}{Roshni~G. Iyer}, \bibinfo{person}{Thuy Vu},
  \bibinfo{person}{Alessandro Moschitti}, {and} \bibinfo{person}{Yizhou Sun}.}
  \bibinfo{year}{2022}\natexlab{}.
\newblock \showarticletitle{Question-Answer Sentence Graph for Joint Modeling
  Answer Selection}.
\newblock \bibinfo{journal}{\emph{arXiv preprint arXiv:2203.03549}}
  (\bibinfo{year}{2022}).
\newblock


\bibitem[Iyer et~al\mbox{.}(2021)]%
        {ba-gnn}
\bibfield{author}{\bibinfo{person}{Roshni~G. Iyer}, \bibinfo{person}{Wei Wang},
  {and} \bibinfo{person}{Yizhou Sun}.} \bibinfo{year}{2021}\natexlab{}.
\newblock \showarticletitle{Bi-Level Attention Graph Neural Networks}.
\newblock \bibinfo{journal}{\emph{ICDM}} (\bibinfo{year}{2021}).
\newblock


\bibitem[Kolyvakis et~al\mbox{.}(2019)]%
        {hyperkg}
\bibfield{author}{\bibinfo{person}{Prodromos Kolyvakis},
  \bibinfo{person}{Alexandros Kalousis}, {and} \bibinfo{person}{Dimitris
  Kiritsis}.} \bibinfo{year}{2019}\natexlab{}.
\newblock \showarticletitle{HyperKG: Hyperbolic Knowledge Graph Embeddings for
  Knowledge Base Completion}.
\newblock \bibinfo{journal}{\emph{arXiv preprint arXiv:1908.04895}}
  (\bibinfo{year}{2019}).
\newblock


\bibitem[Lehmann et~al\mbox{.}(2015)]%
        {db}
\bibfield{author}{\bibinfo{person}{Jens Lehmann}, \bibinfo{person}{Robert
  Isele}, \bibinfo{person}{Max Jakob}, \bibinfo{person}{Anja Jentzsch},
  \bibinfo{person}{Dimitris Kontokostas}, \bibinfo{person}{Pablo~N. Mendesf},
  \bibinfo{person}{Sebastian Hellmann}, \bibinfo{person}{Mohamed Morsey},
  \bibinfo{person}{Patrick van Kleef}, \bibinfo{person}{Soren Auer}, {and}
  \bibinfo{person}{Christian Bizer}.} \bibinfo{year}{2015}\natexlab{}.
\newblock \showarticletitle{DBpedia: A large-scale, multilingual knowledge base
  extracted from Wikipedia}. In \bibinfo{booktitle}{\emph{Semantic Web
  Journal}}. \bibinfo{pages}{167--195}.
\newblock


\bibitem[Meng et~al\mbox{.}(2019)]%
        {spherical-dist}
\bibfield{author}{\bibinfo{person}{Yu Meng}, \bibinfo{person}{Jiaxin Huang},
  \bibinfo{person}{Guangyuan Wang}, \bibinfo{person}{Chao Zhang},
  \bibinfo{person}{Honglei Zhuang}, \bibinfo{person}{Lance Kaplan}, {and}
  \bibinfo{person}{Jiawei Han}.} \bibinfo{year}{2019}\natexlab{}.
\newblock \showarticletitle{Spherical Text Embedding}.
\newblock \bibinfo{journal}{\emph{NeurIPS}} (\bibinfo{year}{2019}).
\newblock


\bibitem[Nickel and Kiela(2017)]%
        {hyperbolic-dist}
\bibfield{author}{\bibinfo{person}{Maximilian Nickel} {and}
  \bibinfo{person}{Douwe Kiela}.} \bibinfo{year}{2017}\natexlab{}.
\newblock \showarticletitle{Poincaré Embeddings for Learning Hierarchical
  Representations}.
\newblock \bibinfo{journal}{\emph{NeurIPS}} (\bibinfo{year}{2017}).
\newblock


\bibitem[Petersen et~al\mbox{.}(2006)]%
        {reim-geom}
\bibfield{author}{\bibinfo{person}{Peter Petersen}, \bibinfo{person}{S. Axler},
  {and} \bibinfo{person}{K.A. Ribet}.} \bibinfo{year}{2006}\natexlab{}.
\newblock \bibinfo{booktitle}{\emph{Riemannian Geometry}}.
\newblock \bibinfo{publisher}{Springer}.
\newblock


\bibitem[Suchanek et~al\mbox{.}(2007)]%
        {yago}
\bibfield{author}{\bibinfo{person}{Fabian~M. Suchanek},
  \bibinfo{person}{Gjergji Kasneci}, {and} \bibinfo{person}{Gerhard Weikum}.}
  \bibinfo{year}{2007}\natexlab{}.
\newblock \showarticletitle{Yago: A Core of Semantic Knowledge}.
\newblock \bibinfo{journal}{\emph{WWW}}.
\newblock


\bibitem[Sun et~al\mbox{.}(2022)]%
        {sun2021self}
\bibfield{author}{\bibinfo{person}{Li Sun}, \bibinfo{person}{Zhongbao Zhang},
  \bibinfo{person}{Junda Ye}, \bibinfo{person}{Hao Peng},
  \bibinfo{person}{Jiawei Zhang}, \bibinfo{person}{Sen Su}, {and}
  \bibinfo{person}{Philip~S. Yu}.} \bibinfo{year}{2022}\natexlab{}.
\newblock \showarticletitle{A Self-supervised Mixed-curvature Graph Neural
  Network}.
\newblock \bibinfo{journal}{\emph{AAAI}} (\bibinfo{year}{2022}).
\newblock


\bibitem[Sun et~al\mbox{.}(2020)]%
        {hyperka}
\bibfield{author}{\bibinfo{person}{Zequn Sun}, \bibinfo{person}{Muhao Chen},
  \bibinfo{person}{Wei Hu}, \bibinfo{person}{Chengming Wang},
  \bibinfo{person}{Jian Dai}, {and} \bibinfo{person}{Wei Zhang}.}
  \bibinfo{year}{2020}\natexlab{}.
\newblock \showarticletitle{Knowledge Association with Hyperbolic Knowledge
  Graph Embeddings}.
\newblock \bibinfo{journal}{\emph{ACL}} (\bibinfo{year}{2020}).
\newblock


\bibitem[Sun et~al\mbox{.}(2019)]%
        {rotatE}
\bibfield{author}{\bibinfo{person}{Zhiqing Sun}, \bibinfo{person}{Zhi-Hong
  Deng}, \bibinfo{person}{Jian-Yun Nie}, {and} \bibinfo{person}{Jian Tang}.}
  \bibinfo{year}{2019}\natexlab{}.
\newblock \showarticletitle{RotatE: Knowledge Graph Embedding by Relational
  Rotation in Complex Space}.
\newblock \bibinfo{journal}{\emph{ICLR}} (\bibinfo{year}{2019}).
\newblock


\bibitem[Toutanova and Chen(2015)]%
        {FB15k-237}
\bibfield{author}{\bibinfo{person}{Kristina Toutanova} {and}
  \bibinfo{person}{Danqi Chen}.} \bibinfo{year}{2015}\natexlab{}.
\newblock \showarticletitle{Observed versus latent features for knowledge base
  and text inference}. In \bibinfo{booktitle}{\emph{Proceedings of the 3rd
  Workshop on Continuous Vector Space Models and their Compositionality}}.
\newblock


\bibitem[Trouillon et~al\mbox{.}(2016)]%
        {complex}
\bibfield{author}{\bibinfo{person}{Théo Trouillon}, \bibinfo{person}{Johannes
  Welbl}, \bibinfo{person}{Sebastian Riedel}, \bibinfo{person}{Éric Gaussier},
  {and} \bibinfo{person}{Guillaume Bouchard}.} \bibinfo{year}{2016}\natexlab{}.
\newblock \showarticletitle{Complex Embeddings for Simple Link Prediction}.
\newblock \bibinfo{journal}{\emph{ICML}} (\bibinfo{year}{2016}).
\newblock


\bibitem[Wang et~al\mbox{.}(2019)]%
        {kg-rec-system}
\bibfield{author}{\bibinfo{person}{Hongwei Wang}, \bibinfo{person}{Fuzheng
  Zhang}, \bibinfo{person}{Jialin Wang}, \bibinfo{person}{Miao Zhao},
  \bibinfo{person}{Wenjie Li}, \bibinfo{person}{Xing Xie}, {and}
  \bibinfo{person}{Minyi Guo}.} \bibinfo{year}{2019}\natexlab{}.
\newblock \showarticletitle{RippleNet: Propagating User Preferences on the
  Knowledge Graph for Recommender Systems}.
\newblock \bibinfo{journal}{\emph{WSDM}} (\bibinfo{year}{2019}).
\newblock


\bibitem[Wang et~al\mbox{.}(2021)]%
        {m2gnn}
\bibfield{author}{\bibinfo{person}{Shen Wang}, \bibinfo{person}{Xiaokai Wei},
  \bibinfo{person}{Cícero Nogueira~dos Santos}, \bibinfo{person}{Zhiguo Wang},
  \bibinfo{person}{Ramesh Nallapati}, \bibinfo{person}{Andrew Arnold},
  \bibinfo{person}{Bing Xiang}, \bibinfo{person}{Philip~S. Yu}, {and}
  \bibinfo{person}{Isabel~F Cruz}.} \bibinfo{year}{2021}\natexlab{}.
\newblock \showarticletitle{Mixed-Curvature Multi-Relational Graph Neural
  Network for Knowledge Graph Completion}.
\newblock \bibinfo{journal}{\emph{WWW}} (\bibinfo{year}{2021}).
\newblock


\bibitem[Yang et~al\mbox{.}(2015)]%
        {distmult}
\bibfield{author}{\bibinfo{person}{Bishan Yang}, \bibinfo{person}{Wen-tau Yih},
  \bibinfo{person}{Xiaodong He}, \bibinfo{person}{Jianfeng Gao}, {and}
  \bibinfo{person}{Li Deng}.} \bibinfo{year}{2015}\natexlab{}.
\newblock \showarticletitle{Embedding Entities and Relations for Learning and
  Inference in Knowledge Bases}.
\newblock \bibinfo{journal}{\emph{ICLR}} (\bibinfo{year}{2015}).
\newblock


\bibitem[Zhang et~al\mbox{.}(2021)]%
        {zhang2021switch}
\bibfield{author}{\bibinfo{person}{Shuai Zhang}, \bibinfo{person}{Yi Tay},
  \bibinfo{person}{Wenqi Jiang}, \bibinfo{person}{Da-cheng Juan}, {and}
  \bibinfo{person}{Ce Zhang}.} \bibinfo{year}{2021}\natexlab{}.
\newblock \showarticletitle{Switch spaces: Learning product spaces with sparse
  gating}.
\newblock \bibinfo{journal}{\emph{arXiv preprint arXiv:2102.08688}}
  (\bibinfo{year}{2021}).
\newblock


\bibitem[Zhang et~al\mbox{.}(2020)]%
        {hake}
\bibfield{author}{\bibinfo{person}{Zhanqiu Zhang}, \bibinfo{person}{Jianyu
  Cai}, \bibinfo{person}{Yongdong Zhang}, {and} \bibinfo{person}{Jie Wang}.}
  \bibinfo{year}{2020}\natexlab{}.
\newblock \showarticletitle{Learning Hierarchy-Aware Knowledge Graph Embeddings
  for Link Prediction}.
\newblock \bibinfo{journal}{\emph{AAAI}} (\bibinfo{year}{2020}).
\newblock


\end{thebibliography}

\appendix

\vspace{-1mm}
\section{General Polar-Cartesian Conversion Framework}
\label{pc-framework}
In this section, we detail our designed general procedures for efficiently converting arbitrary $d$-dimensional embeddings between the polar and Cartesian coordinate representations, which are utilized by sub-models in \ourmodel. These procedures include: (1) the Cartesian to polar coordinate transformation (MCP) and (2) the polar to Cartesian coordinate transformation (MPC). 
\vspace{-3mm}
\begin{figure}[h]
    \centering
    \includegraphics[width=0.95\linewidth]{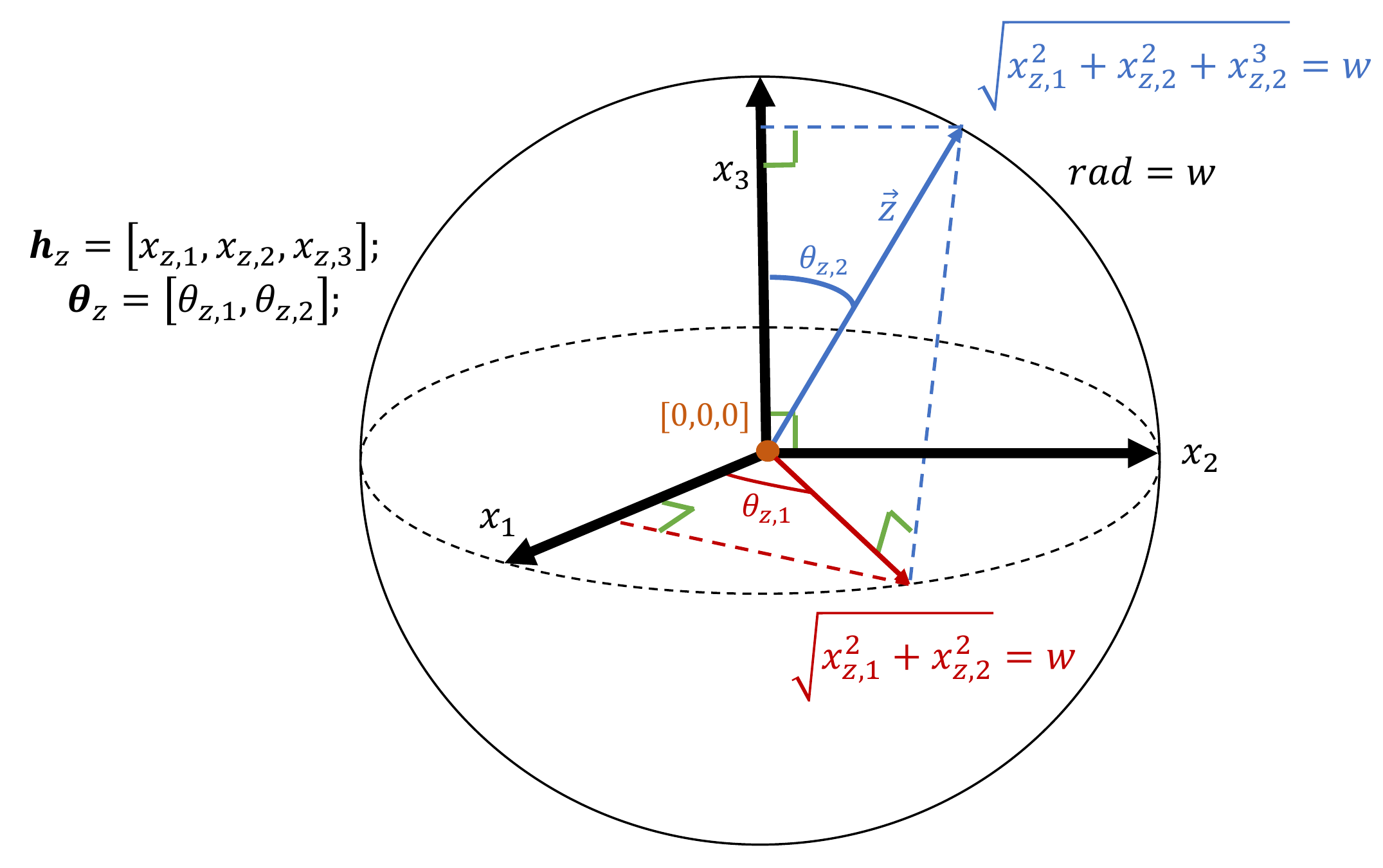}
    \caption{\textmd{Illustration of vector $\boldsymbol{z}$ in spherical space on 3-dimensional surface. Angular dimensions of $\boldsymbol{z}$ are $\theta_{z,1}$, which is the distance from the $x_{1}$-axis to $\boldsymbol{z}$ projected onto the $x_{1}$-$x_{2}$ plane, and $\theta_{z,2}$, which is the distance from the $x_{3}$-axis to $\boldsymbol{z}$ projected onto the $x_{2}$-$x_{3}$ plane.}}
    \label{fig:mpc-mcp-ex}
    \vspace{-4mm}
\end{figure}

\subsection{MCP Transformation Procedure.}
\label{sec-mcp}
Using Figure~\ref{fig:mpc-mcp-ex}, we derive transformations for the 3-dimensional spherical surface, and extend to a $d$-dimensional spherical surface:
\begin{align*}
    \mathrm{tan}(\theta_{z,1}) & = \frac{x_{z,1}}{x_{z,2}}; \mathit{rad} = w \\
    \mathrm{cot}(\theta_{z,2}) & = \frac{x_{z,3}}{\sqrt{x_{z,1}^{2} + x_{z,2}^{2}}}; \mathit{rad} = w \\
    \mathrm{cot}(\theta_{z,3}) & = \frac{x_{z,4}}{\sqrt{x_{z,1}^{2} + x_{z,2}^{2} + x_{z,3}^{2}}}; \mathit{rad} = w \\
    & \vdots \\
    \mathrm{cot}(\theta_{z,d-1}) & = \frac{x_{z,d}}{\sqrt{x_{z,1}^{2} + x_{z,2}^{2} + ...+ x_{z,d-1}^{2}}}; \mathit{rad} = w\\
    \implies \\
    \theta_{z,1} & = \mathrm{tan}^{-1}(\frac{x_{z,2}}{x_{z,1}}); \mathit{rad} = w\\
    \theta_{z,i} & = \mathrm{cot}^{-1}{\frac{x_{z,i+1}}{\sqrt{\sum_{i'=1}^{i}{x_{z,i'}^{2}}}}}, i \in [2, d-1]; \mathit{rad} = w\\
\end{align*}

\begin{figure*}[h]
    \centering
    \includegraphics[width=0.8\linewidth]{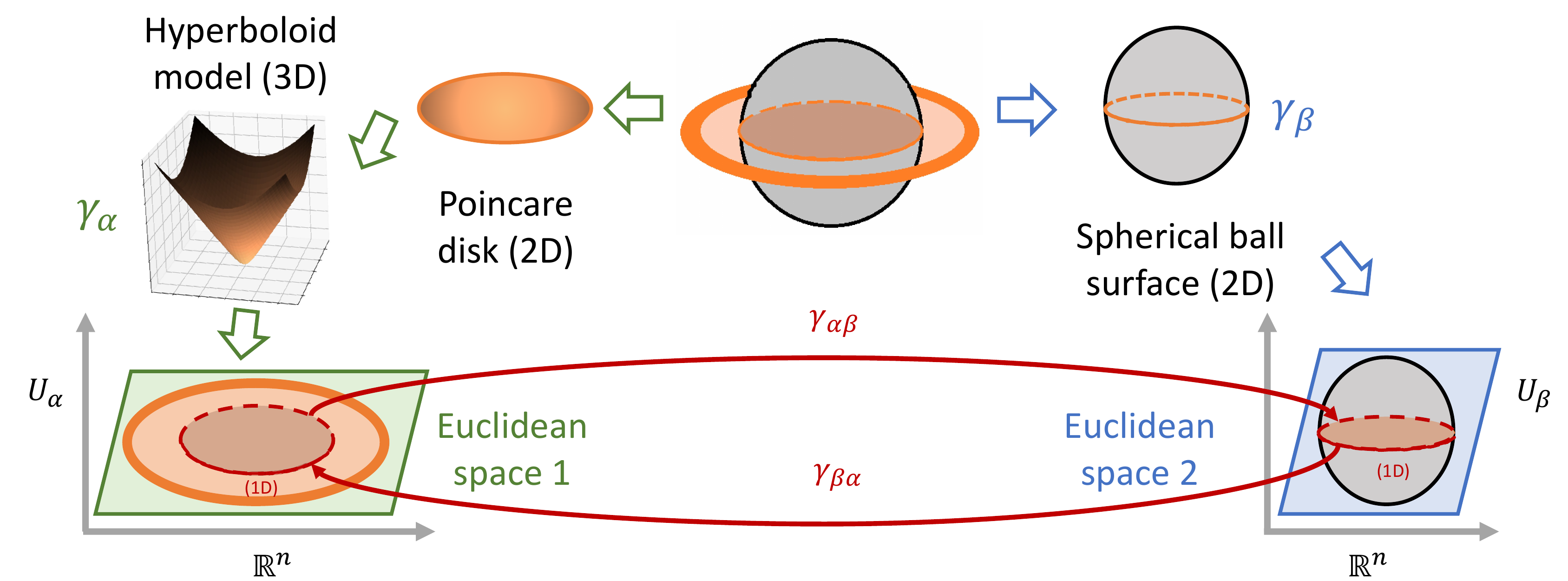}
    \caption{\textmd{Riemannian retraction operation for the bridge space of \ourmodel}}
    \label{fig:retraction}
    \vspace{-3mm}
\end{figure*}

\subsection{MPC Transformation Procedure.}
\label{sec-mpc}
Using Figure~\ref{fig:mpc-mcp-ex}, we derive transformations for the $d$-dimensional spherical surface, as follows with:
\begin{align*}
    x_{z,1} & = \mathrm{cos}(\theta_{z,1}); \mathit{rad} = w\\
    x_{z,d-i} & = \big(\prod_{i'=d-i}^{d-1}\mathrm{sin}(\theta_{z,i'})\big) \cdot \mathrm{sin}(\theta_{z,d-i-1}); \mathit{rad} = w\\
    x_{z,d} & = \mathrm{cos}(\theta_{z,d-1}); \mathit{rad} = w\\
\end{align*}

\paragraph{\normalfont{\textbf{Theorem.}}} The MCP procedure and MPC procedure are closed in the spherical space. 

\begin{proof}
The above equations for the MCP and MPC procedures directly use properties of trigonometry for right triangles for each embedding dimension $\theta_{i}, i \in [1, d-1]$, ensuring that the transformation is closed. 
\end{proof}

\section{Batch Vector Transformation Procedure}
\label{batch-vec-trans}

We illustrate the batch version of the rotation operation presented in Section~\ref{subsec:inst-modeling}. Consider for entity $e_{i}$, there are $K$ relations connected to $e_{i}$. Instead of applying $f_{\mathrm{rot}}$ to each one of the $K$ relations, we can perform the following to compute the result efficiently:
{
\begin{gather*}
\renewcommand\arraystretch{1.3}
f_{\mathrm{rot}} 
    \mleft(
    \boldsymbol{h}_{e_{i}}^{S},
    \mleft[
    \begin{array}{c}
      \boldsymbol{\theta}_{r_{1}}^{S} \\
      \vdots \\
      \boldsymbol{\theta}_{r_{K}}^{S}
    \end{array}
    \mright]
\mright) 
= 
    \mleft[
    \begin{array}{c}
      ({\theta}_{e_{i},1}^{S}+{\theta}_{r_{1},1}^{S}) + \dots + ({\theta}_{e_{i},d}^{S}+{\theta}_{r_{1},d}^{S}) \\
      \vdots \\
      ({\theta}_{e_{i},1}^{S}+{\theta}_{r_{K},1}^{S}) + \dots + ({\theta}_{e_{i},d}^{S}+{\theta}_{r_{K},d}^{S})
    \end{array}
    \mright] \mathrm{mod} 2\pi \\
=
    \mleft[
    \begin{array}{c}
      f_{\mathrm{rot}}(\boldsymbol{h}_{e_{i}}^{S},\boldsymbol{\theta}_{r_{1}}^{S}) \\
      \vdots \\
      f_{\mathrm{rot}}(\boldsymbol{h}_{e_{i}}^{S},\boldsymbol{\theta}_{r_{K}}^{S})
    \end{array}
    \mright]
\end{gather*}

}

\paragraph{\normalfont{\textbf{Theorem.}}} The batch vector transformation procedure is closed.

\begin{proof}
Each relation has $f_{\mathrm{rot}}(\cdot)$ applied to the entity, and $f_{\mathrm{rot}}(\cdot)$ has been shown to be closed because every dimension of the transformed head is within the angular constraints e.g., by $\mathrm{mod} 2\pi$, and norm space constraints because  $f_{\mathrm{rot}}(\cdot)$ preserves the norm space of the spherical space.
\end{proof}

\section{Riemannian retraction operation for the bridge space of \ourmodel} Figure~\ref{fig:retraction} describes the Riemannian retraction operation $\mathcal{R}$ for the bridge space to the tangent Euclidean space. The bridge space is formed by intersecting the surface of the spherical ball with the Poincaré disk in the hyperbolic space. As shown in the figure, we separately map the Poincaré disk and the spherical ball to different tangent Euclidean spaces, and enable for joint communication between the two spaces to represent the bridge entities. On the left, we perform isomorphic mapping from the Poincaré disk, $\mathcal{H}^{d}$, to the hyperboloid model, $\mathcal{B}^{d+1}$ as in~\cite{hgcn} using:
\begin{equation}
\mathcal{R}_{\mathcal{H}^{d} \rightarrow \mathcal{B}^{d+1}}(x_{1},...,x_{d}) = \frac{2x_{1},...,2x_{d}, 1+\norm{\textbf{x}}_{2}^{2}}{1 - \norm{\textbf{x}}_{2}^{2}} \end{equation}
\begin{equation}
\mathcal{R}_{\mathcal{B}^{d+1} \rightarrow \mathcal{H}^{d}}(x_{1},...,x_{d+1}) = \frac{(x_{1}, ..., x_{d})}{x_{d+1}+1} \end{equation}
, then using logarithmic mapping map to the tangent Euclidean space 1. On the right, we use logarithmic mapping to map the spherical ball to the tangent Euclidean space. Then we allow for transformation between the two Euclidean spaces through transformation mappings $\gamma_{\alpha\beta}$ and $\gamma_{\beta\alpha}$ such that 
\begin{equation}
    \gamma_{\alpha\beta} = \gamma_{\beta} \circ \gamma_{\alpha}^{-1}
\end{equation}
\begin{equation}
    \gamma_{\beta\alpha} = \gamma_{\alpha} \circ \gamma_{\beta}^{-1}
\end{equation}









\end{document}